\documentclass{article}

\usepackage[preprint]{neurips_2024}

\usepackage[dvipsnames]{xcolor}
\usepackage[utf8]{inputenc} % allow utf-8 input
\usepackage[T1]{fontenc}    % use 8-bit T1 fonts
\usepackage[frozencache,cachedir=.]{minted}
\usepackage{graphicx}
\usepackage{wrapfig}
\usepackage[colorlinks=true,citecolor=blue,linkcolor=blue]{hyperref}
\usepackage{url}
\usepackage{booktabs}
\usepackage{caption}
\usepackage{subcaption}
\usepackage{enumitem}
\usepackage{natbib}
\usepackage{amsmath}
\usepackage{amssymb}
\usepackage{svg}
\usepackage{pdfpages}
\usepackage{graphicx,subcaption}

\usepackage{xspace}
\newcommand{\eg}{\emph{e.g.}\xspace} 
 
\newcommand{\ie}{\emph{i.e.}\xspace} 
 
\newcommand{\cf}{\emph{cf.}\xspace}

\newcommand{\interpreter}[0]{{\small INTERPRETER}\xspace}

\usepackage[linesnumbered,ruled,vlined]{algorithm2e}
\DontPrintSemicolon

\SetKwComment{Comment}{\color{green!50!black}// }{}

\newcommand{\assign}{\leftarrow}
\SetKwProg{Function}{function}{}{}

\title{Interpretable and Editable Programmatic Tree Policies for Reinforcement Learning}

\author{%
Hector Kohler$^{1*}$ \quad Quentin Delfosse$^{2,3}$\thanks{Equal contributions.} \quad Riad Akrour$^1$ \\
\textbf{Kristian Kersting}$^{2,4}$ \quad \textbf{Philippe Preux}$^1$\\
$^1$ Universit\'e de Lille, Inria \quad $^2$TU Darmstadt \quad $^3$ ATHENE \quad $^4$ Hessian AI, DFKI\\
\texttt{\{hector.kohler,riad.akrour,philippe.preux\}@inria.fr}\\
\texttt{\{quentin.delfosse,kersting\}@cs.tu-darmstadt.fr}\\
}
\begin{document}

\maketitle

\begin{abstract}
Deep reinforcement learning agents are prone to goal misalignments. The black-box nature of their policies hinders the detection and correction of such misalignments, and the trust necessary for real-world deployment. 
So far, solutions learning interpretable policies are inefficient or require many human priors.
We propose \interpreter\footnote{\url{https://anonymous.4open.science/r/tuto-interpreter-C56E/tuto-interpreter.ipynb}}, a fast distillation method producing INTerpretable Editable tRee Programs for ReinforcEmenT lEaRning.
We empirically demonstrate that \interpreter compact tree programs match oracles across a diverse set of sequential decision tasks and evaluate the impact of our design choices on interpretability and performances. 
We show that our policies can be interpreted and edited to correct misalignments on Atari games and to explain real farming strategies. 
\end{abstract}

\section{Introduction}

\textbf{Why interpretability?}
The last decade has seen a surge in the performance of machine learning models, in supervised learning~\citep{AlexNet,attention} and in reinforcement learning (RL)~\citep{Mnih2015dqn,Schulman2017ProximalPO,bhatt2024crossq}. These achievements rely on deep neural networks that are often described as black-box models~\citep{Murdoch, Guidotti18, Arrieta}, trading interpretability for performance. In many real world tasks, predictive models can hide undesirable biases, such as the ones listed by \citet{Guidotti18}, hindering trustworthiness towards AI agents. Gaining trust is one of the main goals of interpretability %~(see Sec.\@ 2.4 of \
\citep{Arrieta}, % for a literature review) 
along with informativeness requests, \ie the ability for a model to provide information on why and how decisions are taken. 
The computational complexity of such informativeness requests can be measured objectively, and \cite{Barcelo20} showed that multi-layer neural networks cannot answer these requests, at least not in polynomial time, whereas explicit structures, \eg, decision trees can.

\textbf{Interpretable RL using transparent models as policies.}
In addition to trust and informativeness, the importance of interpretability is further highlighted in RL for addressing shortcut learning, where agents learn to exploit spurious correlations instead of mastering the intended tasks. 
This leads to poor generalization, often observed as goal misgeneralization in deep RL~\citep{Langosco2022goal}. 
Explainable methods, \eg importance maps, have been used to pinpoint such flawed strategies~\citep{SchramowskiSTBH20,RasXGD22,RoyKR22,SaeedO23}. 
However, while these methods reveal which inputs affect decisions, they do not clarify \textit{how} they are used within the decision-making process, which is necessary for detecting and correcting misalignments.
~\cite{delfosse2024interpretable} show that using interpretable concepts as policy states (\eg\@ extracted objects instead of pixels), and transparent policies, (\ie for which the input transformation can be followed,~\citep{milani2022survey,glanois2022xrl}) ease the detection and correction of such misalignment, as transparent policies allow experts interventions.

\begin{figure}[t!]
\centering
    \includegraphics[width=\textwidth]{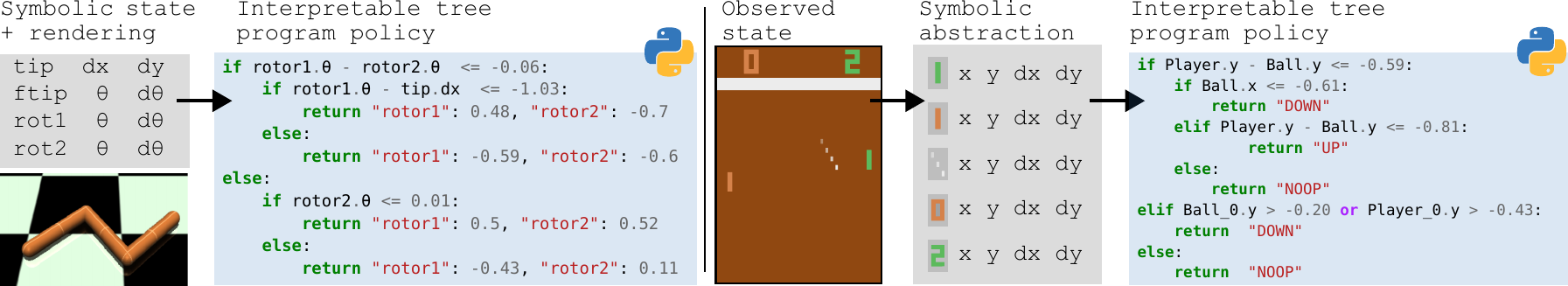}
    \caption{\textbf{\interpreter provides editable interpretable policy}, as a Python tree programs, illustrated on the Swimmer (left) and Pong (right) environments.}
    \label{fig:motiv}
\end{figure}

% \textbf{How to achieve interpretability?} RL algoritms that can provide transparent explanations decisions fall in three categories: i) algorithms that output a transparent policy by imitating a black-box oracle defined over interpretable states \citep{VermaMSKC18,Bastani2018VerifiableRL,landajuela,akrour21}, ii) algorithms that learn from scratch a transparent policy defined over interpretable states \citep{IBMDP}, but they sometimes require carefully human-designed state relations primitives or low-level policy primitives \citep{Delfosse2023InterpretableAE,qiu2022programmatic}, iii) finally algorithms that learn from scratch both an interpretable state space and a transparent policy such as the recent INSIGHT \citep{Luo2024INSIGHTEN} but makes use of a computationally heavy large language model (LLM) to explain their transparent policy.
\textbf{Transparency is not enough.} Having an algorithm learning a transparent policy is not enough to achieve interpretability. Imitation-based solutions like \citep{VermaMSKC18,Bastani2018VerifiableRL,landajuela} might only require an oracle but return either transparent policies with too many decision rules to be considered interpretable \citep{Bastani2018VerifiableRL} or are only tested on small toy problems \citep{VermaMSKC18,landajuela,IBMDP}. On the other hand, \cite{Delfosse2023InterpretableAE,delfosse2024interpretable} outputs transparent policies for various tasks but do require carefully designed human primitives. 
Another approach is to search for a transparent policy among human-designed polynomial equations with deep RL, and use a large language model (LLM) a posterirori to explain the equations~\citep{Luo2024INSIGHTEN}, but such explanations are more likely to emerge from the LLM's acquired knowledge of the RL tasks rather than its ability to understand polynomials.

%, ii) algorithms that learn from scratch a transparent policy defined over interpretable states \citep{IBMDP}, but they sometimes require carefully human-designed state relations primitives or low-level policy primitives \citep{Delfosse2023InterpretableAE,qiu2022programmatic}, iii) finally algorithms that learn from scratch both an interpretable state space and a transparent policy such as the recent INSIGHT \citep{Luo2024INSIGHTEN} but makes use of a computationally heavy large language model (LLM) to explain their transparent policy.

\textbf{Contributions.}  
%In this work, we focus on algorithms %from the first category
%based on imitation learning and policy distillation. In particular, we study variants of imitation learning algorithms like Dagger and VIPER \citep{Ross2010ARO,Bastani2018VerifiableRL} that are fast and simple to implement.
% Because transparent policies are interpretable only if they employ a limited number of decision rules and concepts used within these rules is limited, to achieve interpretability over explainability programmatic policies complexity needs to be regularized. This can impact oracle matching performances of such imitation learning algorithms. In this work, we introduce \interpreter, a policy distillation algorithm fitting regularized oblique trees \citep{murthy1994system} to neural oracle such as DQN and PPO agents \citep{Mnih2015dqn,Schulman2017ProximalPO} and converting them to editable Python programs. We first summarize our contributions then we provide the scientific background of interpretable reinforcement learning:
%Because 
In this work, we introduce \interpreter, a policy distillation method that extracts compact editable tree programs (\cf Figure~\ref{fig:motiv}). Our algorithm fits regularized oblique trees~\citep{murthy1994system} to neural oracle such as DQN and PPO agents \citep{Mnih2015dqn,Schulman2017ProximalPO} and convert them to editable Python programs. Specifically, our contributions are:
\begin{enumerate}[leftmargin=0.7cm]
    % \item In section \ref{sec:performances}, we show that \interpreter outputs compact programs that match neural oracles for various MDPs in few minutes.
    \item We introduce \interpreter, that extracts, without human priors, compact editable tree programs matching oracles for various RL tasks in few minutes.
    % \item In section \ref{sec:algo-abl}, we justify using Dagger imitation over $Q$-Dagger and oblique trees over axes-parallel trees with an ablation study.
    \item We conduct ablation studies to identify the responsibility of \interpreter's components on the extracted policies' performances.
    \item We show that \interpreter tree programs can be interpreted and edited by human experts.
    % \item We convert show Python programs and make a qualitative study of this choice of interpretable policy representation.
\end{enumerate}
We now introduce the background on interpretable reinforcement learning.

\section{Background and notations}\label{sec:background}
\textbf{A Markov decision process} (MDP) $\mathcal{M}$ is a tuple $\langle S, A, R, P, \gamma \rangle$ \citep{puterman2014markov}. We consider continuous states $S \subsetneq \mathbf{R}^p$ and discrete or continuous actions ($\dim(A) \geq 1$). $R: S \times A \rightarrow \mathbb{R}$ is the scalar reward function; and $P: S \times A \rightarrow \Delta S$ are the transition probabilities ($p(s_{t+1}=s|a_t=a)\sim P$). A discrete (resp. stochastic) policy is a mapping $\pi:S\rightarrow A$ (resp. $\pi:S\rightarrow \Delta A$). A policy takes actions in an MDP through time and receives rewards $R(s_t, \pi(s_t))$. Given a policy $\pi$, the value of $\pi$ in the state $s$ after taking action $a$ is the expected discounted cumulative reward: $Q^{\pi}(s,a)=R(s,a) + \gamma \mathbb{E}_{s'\sim P}[Q^{\pi}(s', \pi(s')]$, with $0<\gamma<1$. 

\textbf{Reinforcement learning} algorithms \citep{sutton2018reinforcement} look either for the optimal $Q$-function $Q^*$ ($Q^*(s,a) \geq Q(s,a)$ for any $Q,s,a$) \citep{Mnih2013PlayingAW, Mnih2015dqn, vanHasseltGS16ddqn}; or for the optimal policy $\pi^*=\operatorname{argmax}_{\pi}\:\mathbb{E}[\sum_t\gamma^tR(s_t, \pi(s_t))]$ \citep{Schulman2017ProximalPO,pmlr-v80-haarnoja18b}. Often we have $Q^*=Q^{\pi^*}$ and $\pi^*(s)=\operatorname{argmax}_a\:Q^*(s,a)$. Our goal is to find an interpretable policy $T^*$ whose performances are close to or equal to $\pi^*$. We first use reinforcement learning to get $\pi^*$ and/or $Q^{\pi^*}$ and then consider two imitation learning methods to find $T^*$. 

\textbf{Imitation learning} transforms a reinforcement learning task into a sequence of supervised learning ones. At each iteration $i$ of Dagger \citep{Ross2010ARO}, $T_i$ is fitted to a dataset of states collected with $T_{i-1}$ and actions given by $\pi^*$ on those states \citep{Ross2010ARO}. $Q$-Dagger \citep{Bastani2018VerifiableRL} further re-weights state-action samples proportional to  $\mathbb{E}_{a \in A}Q^{\pi^*}(s,a) - \operatorname{min}_{a \in A}Q^{\pi^*}(s,a)$. 
\newpage

\begin{figure}[t!]
\centering
    \includegraphics[width=0.85\textwidth]{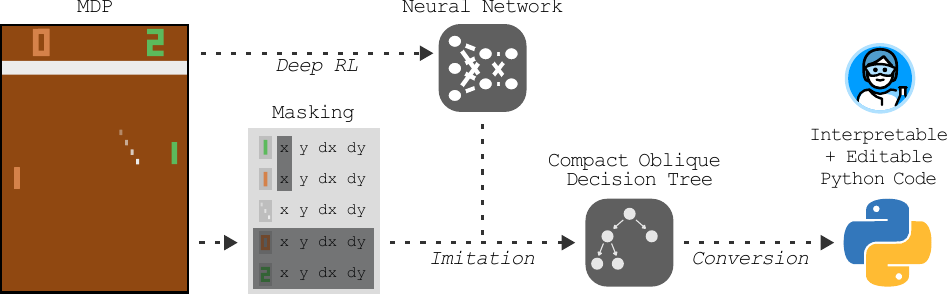}
    \caption{\textbf{\interpreter's Distillation process.} The MDP state-action space is simplified (idle features and equivalent actions are masked), then an oblique tree policy imitates the oracle. Finally, the policy is then translated to readable and executable code: experts can verify and edit.}
    \label{fig:method}
    \vspace{-0.1cm}
\end{figure}

\section{Method}\label{sec:algo}
% \quentin{
% \interpreter produces Python code that solve RL tasks from neurosymbolic states (\cf Figure~\ref{fig:method}, Left). 
% In many RL environments, agents only have access (\eg Atari ones) to raw input images. For these, UNIC can first make use of an object detection method~\citep{delfosse2021moc, zhao2023fast}. 
% Then UNIC trains a neural policy with the symbolic state. This non interpretable policy is then distilled into a Python code, usable to play the game. We hereafter describe this distillation process.
% }{}
\textbf{Imitation Learning routine.} 
To find an interpretable tree program policy $T^*$,
\interpreter uses two different imitation subroutines, depending on the nature of the oracle $\pi^*$. For MDPs with discrete actions, if both the oracle policy $\pi^*$ and the associated state-action value function $Q^{\pi^*}$ are accessible, the oracle data are collected with the $Q$-Dagger subroutine from \citep{Bastani2018VerifiableRL}, described in section \ref{sec:background}) and corresponding to line \ref{alg:ifelse} of algorithm \ref{alg:iqdagger}. On the other hand, if the action space is continuous or only the oracle's policy $\pi^*$ is accessible, the oracle data are collected with the Dagger subroutine (algorithm 1 of \citep{Ross2010ARO}). In later case, \cite{Bastani2018VerifiableRL} still recommend using the $Q$-Dagger routine (over Dagger) with $\log(\pi^*(s,a))$ to reconstruct $Q^{\pi^*}$. %However, we only use $Q$-Dagger over Dagger when actions are discrete and the policy is being greedy w.r.t a $Q$ function. Indeed, 
In practice, \interpreter performances do not depend on the imitation subroutine (\cf experiment~\ref{fig:disagreement}). 
Despite the associated runtime and memory costs for storing neural state-action value functions and performing the forward passes (\cf line \ref{alg:viper2} of algorithm \ref{alg:iqdagger}), we still use $Q$-Dagger in the case above, as it has better guarantee over Dagger when the $Q^{\pi^*}(s, a)$ are well-estimated~\citep{Bastani2018VerifiableRL}.

\begin{figure}[b]
    %\vspace{-0.12cm}
    \centering
    \includegraphics[width=1.\textwidth]{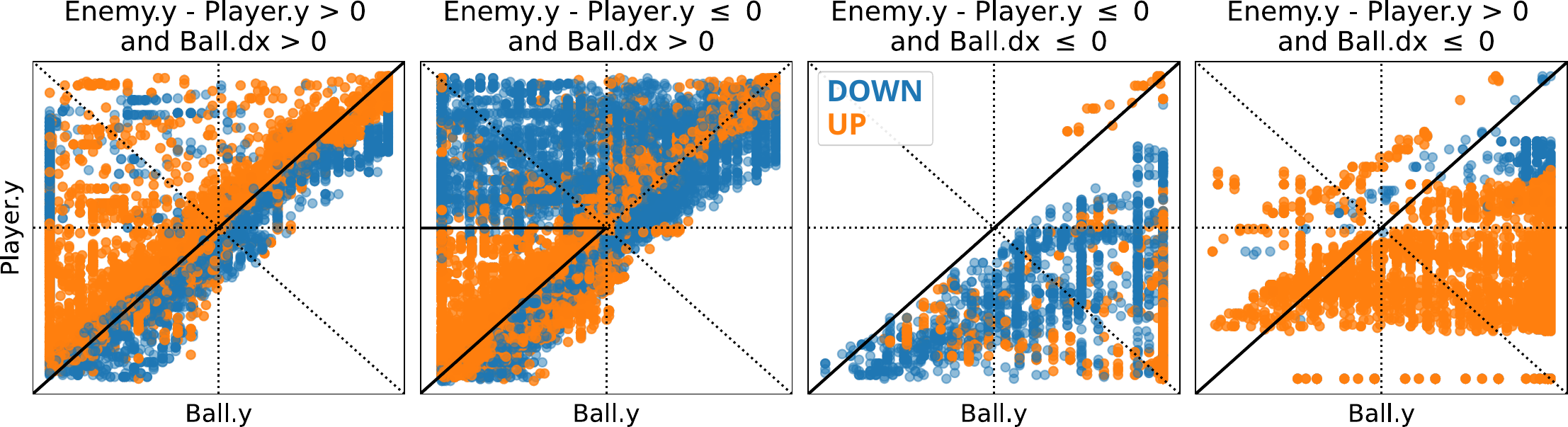}
    %\vspace{3mm}
    \caption{\textbf{Oracle decision rules are oblique} illustrated on PPO for different state space partitions of the Pong environment. Decisions boundaries are both oblique and parallel.}
    \label{fig:pong_states}
\end{figure}

\textbf{Oblique decision trees.} One can imitate oracles with programs that make tests of linear combinations of features. Many oracles learn oblique or more complex decision rules over an MDP state space. This is illustrated in Figure~\ref{fig:pong_states} where a PPO neural oracle creates oblique partitions of the state-space for the Pong environments. Programs that test only individual features would fail to fit this partition (\cf Figure~\ref{fig:pong_states}). 
We thus modify CART~\citep{breiman}, an algorithm returning an axes-parallel trees for regression and supervised classification problems, for it to return oblique decision trees. 
% Essentially, oblique trees trade-off interpretability of axes-parallel trees for better expressivity: each oblique tree node tests a linear combination of features. 
In addition to single feature tests, our oblique trees consider linear combinations of two features with weights $1$ and $-1$, \eg\@ for MDP states $s_i \in \mathbb{R}^{p}$, the oblique features values are $s^{oblique}_i = \{s_{i1} - s_{i0}, s_{i2} - s_{i0}, ..., s_{ip} - s_{i0}, ...,s_{ip-1} - s_{ip}\} \in \mathbb{R}^{p^2}$. For example, using an oracle dataset with $n$ state-actions pairs: $(\bar{S}, \Bar{A} = \pi^{*}(\bar{S})) \subsetneq \mathbb{R}^{n\cdot(p+\dim(A))}$, we obtain oblique decision trees by fitting $(\bar{S},\bar{S}^{oblique} , \Bar{A} = \pi^{*}(\bar{S})) \subsetneq \mathbb{R}^{n\cdot(p(p+1)+\dim(A))}$. 
Given $\bar{S}$, computing $\bar{S}^{oblique}$ can be done efficiently by computing the values of the lower (or upper) triangles in the $\bar{S} \otimes \bar{S} - (\bar{S} \otimes \bar{S})^T$ tensor (excluding the diagonals) (\cf line~\ref{alg:oblique2} of Algorithm~\ref{alg:iqdagger}). We further demonstrate the superiority of oblique trees in our experimental evaluation on a diverse set of RL tasks.

\begin{wraptable}[7]{r}{0.55\textwidth}
\centering
\vspace{-0.5mm}
\caption{ \textbf{\small Automated masking reduces the number of features in MDP}, illustrated on $8$ Atari environments.}
\setlength{\tabcolsep}{3pt}
\scalebox{0.9}{
\begin{tabular}{@{}ccccccccc@{}}
\toprule
MDP        & \small Ast. & Box. & \small Free. & \small Kang. & \small Pong & \small Sea. & \small SpaceI. & Ten. \\ \midrule
Full    & 100         & 8    & 48           & 196          & 12          & 172         & 176            & 16   \\
Simplified & 90          & 8    & 22           & 28           & 8           & 54          & 164            & 16   \\ \bottomrule
\end{tabular}}
\label{tab:repartitions}
\end{wraptable}
\textbf{The complexity} of building the tree (line \ref{alg:cart2} of algorithm \ref{alg:iqdagger}) is $O(pn\log_2(n))$ when no maximum tree depth is given, and $O(pnD)$ with a maximum tree depth of $D$ \citep{complexcart}. 
In particular, at iteration $i$ of \interpreter the complexity of building the tree is $O(p(p+1)itD)$, as rollouts of $t$ MDP transitions are aggregated (line \ref{alg:aggreg}) and oblique features are added to states (line \ref{alg:oblique2}). 
This means that at each iteration $i$, the cost of computing an oblique tree is $p+1$ times the cost of computing an axes-parallel tree. In \interpreter we pass $K$ the maximum number of leaf nodes as an argument. A tree with $K$ leaf nodes has $2K - 1$ total nodes and a depth of at most $D=K-1$.  
% \begin{wrapfigure}[16]{r}{0.53\textwidth}
%     \vskip -0.3cm
%     \centering
%     \tiny
%     \vskip -0.4cm
%     \inputminted{Python}{policy_programs/play_Pong_8_leaves_15.0.py}
%     \vskip -0.2cm
%     \caption{\small Program to play Pong returned by \interpreter.}
% \end{wrapfigure}

\textbf{Conversion into ready-to-use programs.}
After the imitation learning subroutine, \interpreter converts the best evaluated oblique tree in line~\ref{alg:best-tree} into a Python program (\cf line~\ref{alg:tree-to-prog}), such as the ones depicted in Figure~\ref{fig:motiv}.
To do so, we turn the internal nodes of the tree into if-else statements. 
This algorithmic step does not involve any randomness nor learning. This Python based representations of our interpretable policies, unlike a tree plot, \interpreter can easily be edited, as shown in section~\ref{sec:eval}. Finally, we perform pruning to further simplify our programs by merging redundant logical branches. We provide details and examples in Appendix~\ref{app:prog_simpl}.
\vspace{-2mm}
\begin{algorithm}[h!]
    \small
    \SetKwProg{Function}{function}{}{}
  \caption{\interpreter}\label{alg:iqdagger}
  \Function{\interpreter($\pi^*$, $\mathcal{M}$, $K$, $N$, $t$, $Q^{\pi^*}=None$)}{
        \If{IsNotNone($Q^{\pi^*}$) \& IsDiscrete($A$)}{\label{alg:ifelse}
      SampleWeighting() $\assign \mathbb{E}_{a \in A}Q^{\pi^*}(s,a) - \operatorname{min}_{a \in A}Q^{\pi^*}(s,a)$\;
    }
    \Else{
        SampleWeghting() $\assign 1$\;
    }
    $\mathcal{M} \assign$ MaskIdleFeatures($\mathcal{M}$) \;\label{alg:idle}
    $(\bar{S}, \bar{S}^{oblique}, \bar{A},\bar{W}) \assign \emptyset$\;
    \For{$i = 1, 2, ...,N$}{
    \If{$i=1$}{$\bar{S}_i \assign$ rollouts($\pi^*,\:\mathcal{M}$, $t$)\;}
    \Else{$\bar{S}_i \assign$ rollouts($T_{i-1},\:\mathcal{M}$, $t$)\;}
    $\bar{A}_i \assign \pi^*(\bar{S}_i)$\;
    $\bar{S}^{oblique}_i \assign \operatorname{Triangles}(\bar{S}_i \otimes \bar{S}_i - (\bar{S}_i \otimes \bar{S}_i)^T)$\;\label{alg:oblique2}
    $\bar{W}_i\assign$ SampleWeighting($\bar{S}_i, Q^{\pi^*}$)\;\label{alg:viper2}
    $(\bar{S}, \bar{S}^{oblique}, \bar{A},\bar{W} )\assign (\bar{S}, \bar{S}^{oblique},\bar{A}, \bar{W}) \cup (\bar{S}_i, \bar{S}^{oblique}_i,\bar{A}_i, \bar{W}_i)$\;\label{alg:aggreg}
    $\bar{T}_i\assign$ \label{alg:}CART($(\bar{S},\bar{S}^{oblique}, \bar{A}, \bar{W} ), K$)\;\label{alg:cart2}
    }
    
    $T^*$ $\assign$ 
    BestTreeEvaluation($\{\bar{T}_1,..., \bar{T}_N\},\:\mathcal{M}$)\;\label{alg:best-tree}

    Prog $\assign$ TreeToProgram($T^*$)\;\label{alg:tree-to-prog}
    \Return{$T^*$, Prog}
}
\end{algorithm}\\
Let us now evaluate {\small INTERPRETER}'s performances, interpretability, and correction possibilities. 

% \section{How well do \interpreter tree programs match oracles performances?}
\section{Experimental Evaluation}
\label{sec:eval}
In this section, we evaluate the compact tree programs provided by \interpreter. Specifically, our experimental evaluation aims at answering the following research questions: \textbf{(Q1)} Can compact \interpreter tree programs match the oracles performances? \textbf{(Q2)} What are the key design choices of \interpreter? \textbf{(Q3)} Can {\small INTERPRETER}'s programs be interpreted and modified by experts?  

% outputs compact tree programs that can match the performance of a given neural oracle. 

% In particular, we highlight that fitting oblique trees rather than classical axes-parallel trees~\citep{Bastani2018VerifiableRL} increases performances over all tested benchmarks. Then we show that \interpreter trees match neural oracles independently of the imitation learning subroutine choice.  

\textbf{Metrics.}
For \interpreter, we denote the maximum number of leaf nodes, $K$, the number of different fitted trees $N$, and the number of transitions, $t$. Given an MDP $\mathcal{M}$, the oracle policy $\pi^*$, and potentially its Q-value function, $Q^{\pi^*}$, from $\mathcal{M}$ to fit each tree, we report in the value of the cumulative reward of the \interpreter tree programs, normalized by the oracle, and often average it over multiple MDPs that have similar properties, such as $M$ MuJoCo robots or $M$ Atari games.

\textbf{Benchmarks.} We tested \interpreter on a set of common RL benchmarks: classic control, Atari games, MuJoCo robot simulations \citep{todorov12mujoco,Bellemare2012TheAL,pmlr-v37-schulman15,vanHasseltGS16ddqn,Schulman2017ProximalPO,pmlr-v80-haarnoja18b}. In particular, for Atari games, we use the stochastic version where actions have a non-zero probability to be repeated as per the recommendations of \citep{MachadoBTVHB18} for best practices of RL training. We use \texttt{gymnasium} \citep{towers_gymnasium_2023} implementations of those benchmarks: we use \texttt{-v4} version of MuJoCo environments, \texttt{-v5} version of Atari games, and the latest versions of classic control.

\textbf{Object-centric Atari.} To reduce the computational burden, and as object detection is not the core focus of this work, we use the neurosymbolic states efficiently extracted by OCAtari \citep{Delfosse2023OCAtariOA} for \interpreter trees to map neurosymbolic states to actions. 
OCAtari extracts these states from the RAM for each Atari environment, with similar accuracies as other extraction methods \cite{lin2020space,zhao2023fast}. 
These states list the depicted game objects $x,y$-coordinates rather than pixels. For further details, we refer to the authors' original publication. %In OCAtari, each instance of a game object is present in the MDP state as seperate features, \eg in SpaceInvaders even though all the alien game objects have the same properties, there will be a different state feature for each one's coordinates.  

\textbf{Neural oracles.} Most interpretable RL algorithms extract transparent policies from black-box oracles, such as deep neural networks~\citep{Bastani2018VerifiableRL,VermaMSKC18,delfosse2024interpretable}. 
For MuJoCo and classic control, we re-use oracles from \texttt{stable-baselines3} zoo~\citep{rl-zoo3}; the final policies obtained from a SAC agent training~\citep{pmlr-v80-haarnoja18b}, and from DQN~\citep{Mnih2015dqn} and PPO~\citep{Schulman2017ProximalPO} for classic control. For Atari tasks, we PPOs with \texttt{stable-baselines3} \citep{stable-baselines3} and the hyperparameters from \citep{Schulman2017ProximalPO}. The oracles' training curves are depicted in Figure~\ref{fig:summary-oracles}, in the Appendix.

\textbf{\interpreter hyperparameters.} Unless stated otherwise, we do between $3$ and $5$ runs of \interpreter in every experiment. Given an oracle policy $\pi^*$ and optionally its associated $Q^{\pi^*}$, each run fits a total of $N\!=\!10$ trees by aggregating $t\!=\!10^4$ transitions at each iteration. 
We vary the imitation learning subroutines, the fitted tree classes, and the maximum number of nodes allowed in each tree ($2K$ with $K$ the maximum number of leaf nodes passed to \interpreter). We use the \texttt{scikit-learn} \citep{scikit-learn} implementation of the CART decision tree algorithm \citep{breiman} with default hyperparameters and $K$ maximum leaf nodes. All experiments are run on a single \texttt{Intel Core i7-8665U@1.90GHz}. Our code is given in supplementary material.
\subsection{\interpreter performances match the oracle performances (Q1)} \label{sec:performances}
\begin{figure}[b]
    \includegraphics[width=1\textwidth]{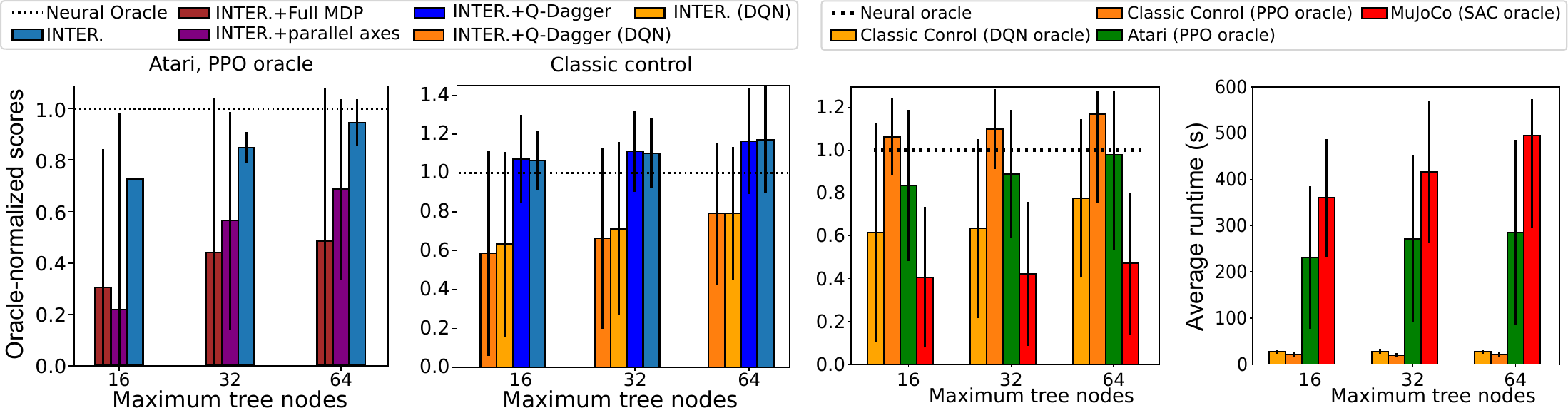}
    \caption{\textbf{\interpreter matches oracles thanks to its design choices.} From left to right: ablated \interpreter, \interpreter with different oracles and imitations, performances and runtimes.}
    \label{fig:ablation}
\end{figure}
% \begin{wrapfigure}[13]{r}{0.55\textwidth}
% \vspace{-0.58cm}
%   \begin{center}
%   \includegraphics[width=.55\textwidth]{figures/iqdagger_atari_mujoco_gym_summary.pdf}
%     \caption{\textbf{\small \interpreter matches oracles:} tree programs performances and \interpreter runtimes}
%     \label{fig:summary-iqdagger}
%     \end{center}
% \end{wrapfigure}
We test \interpreter on the aforementioned benchmarks using algorithm \ref{alg:iqdagger}. 
As shown in Figure~\ref{fig:ablation}, \interpreter tree programs composed of as few as $16$ nodes can outperform their neural oracle on classical control tasks and on Asterix, Pong and SpaceInvaders Atari games (\cf Appendix~\ref{fig:detailed-ablation}). 
In particular, for PPO neural oracles on classic control tasks, \interpreter consistently outperform their neural counterparts. 
In general, \interpreter performances increase with the number of nodes. 
With $64$ nodes, \interpreter programs consistently match or surpass neural oracles' performances on $6$ out of $8$ Atari tasks, and obtain comparable scores for the $2$ others. 
For MuJoCo, \interpreter match oracles for HalfCheetah and Swimmer, but fail at controlling Walker2d and Hopper. 
Importantly, {\small INTERPERTER}'s trees and programs can be extracted within in a few minutes. The greatest runtime bottleneck is MDP rollouts. 
Finally, when the training of neural RL agents fails (\ie has not converged), \eg, DQN curves on Figure~\ref{fig:summary-oracles}, the imitation process is noisy, and the \interpreter trees have poor oracle normalized performances, \cf DQN error bars in Figure~\ref{fig:ablation}. 

\subsection{\interpreter Ablation (Q2)}\label{sec:algo-abl}

\textbf{Imitation learning subroutines.}

\begin{wrapfigure}[10]{r}{0.53\textwidth}
\vspace{-2cm}
  \begin{center}
  \includegraphics[width=0.53\textwidth]{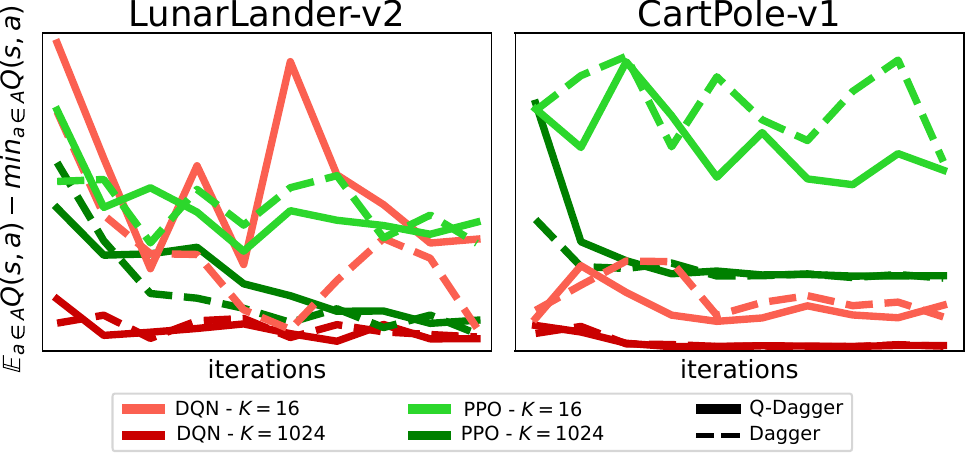}
        \caption{\textbf{\small $Q$-Dagger does not improve sampling,} shown by the similar loss (to Dagger) during the extraction for different oracles and imitations.}
        \label{fig:disagreement}
    \end{center}
\end{wrapfigure}
We use \interpreter to extract decision tree policies for classic control problems, varying the imitation learning subroutine. To do so, we force \interpreter use $Q$-Dagger \citep{Bastani2018VerifiableRL} (\cf line \ref{alg:ifelse} of Algorithm~\ref{alg:iqdagger}), even when the neural oracle is a stochastic policy $\pi^*$ from a PPO agent. In that case, we use $\log\pi^*$ as $Q$-function. 
As depicted in Figure~\ref{fig:disagreement}, for a given oracle and a given maximum number of nodes, $Q$-Dagger and Dagger minimize the $Q$-Dagger loss $\mathbb{E}_{a \in A}Q^{\pi^*}(s,a) - \operatorname{min}_{a\in A}Q^{\pi^*}(s,a)$ similarly. 
\cite{Bastani2018VerifiableRL} show that $Q$-Dagger trees have fewer nodes but are as good as Dagger trees on their own implementation of the Pong environment. 
Our results on the original Pong Atari Learning Environment contradict this claim, as shown in Figure~\ref{fig:detailed-ablation}, where for a given number of nodes both Dagger and $Q$-Dagger trees perform similarly in average. 
This can be due to the fact that either (i) Dagger trees are shorter than $Q$-Dagger trees when the oracle is not a well-estimated $Q$-function and/or (ii) the fitted trees do oblique tests and/or (iii) the nodes regularization is done by bounding the number of leaves rather than the depth.

% \begin{figure}[t]
% % \vspace{-1.6cm}
%   \begin{center}
%   \centering
%      \begin{subfigure}[b]{0.48\textwidth}
%          \centering
%          % \includegraphics[width=0.95\textwidth]{figures/gym_disagreement.pdf}
%          \includegraphics[width=1\textwidth]{figures/iqdagger_atari_mujoco_gym_summary.pdf}
%      \newsubcap{\textbf{\small \interpreter matches oracles:} tree programs performances and runtimes.}
%     \label{fig:summary-iqdagger}
%      \end{subfigure}
%      \hfill
%      \begin{subfigure}[b]{0.48\textwidth}
%          \centering
%          \includegraphics[width=1\textwidth]{figures/gym_disagreement.pdf}
%           \newsubcap{\textbf{\small $Q$-Dagger does not improve sampling,} shown by the similar loss (to Dagger) during the extraction for different oracles and imitations.}
%         \label{fig:disagreement}
%      \end{subfigure}
        
%     \end{center}
% \end{figure}
\textbf{Oblique decision trees.} We compare {\small INTERPRETER}'s tree program performances when fitting classical axes-parallel trees \citep{breiman} -- that have internal nodes such as ``is the x-coordinate of the player $\leq v$?'' -- with fitting oblique trees \citep{murthy1994system} that have internal nodes like ``is the x-coordinate of the player — the previous x-coordinate of the player $\leq v$?''. On Atari games (Figure~\ref{fig:ablation}), using oblique trees over ones for which decisions are parallel to the axes is critical to match oracle performances. As demonstrated in a per-game ablation (\cf Figure~\ref{fig:detailed-ablation} in the Appendix), no axes-parallel tree can get close to the oracle performances, even with a high number of internal nodes. This is supported by our early observation (\cf Figure~\ref{fig:pong_states}).
Further, the linear combinations of input features used in oblique tree programs do not hinder interpretability, as these features are still very human understandable (representing \eg distance over one axis). 
However, for some complicated control problems, such as Tennis (Figure~\ref{fig:detailed-ablation} in the Appendix) or Walker2d (Figure~\ref{fig:summary-oracles}, right), no oblique tree program matches the oracle performances even with $64$ nodes.

\textbf{Removing idle state features.} 
As explained in Section~\ref{sec:algo}, the oblique tree programs' input features is way higher than the one of parallel ones. 
This number will particularly explode in environments such as Kangaroo, Seaquest, and SpaceInvaders, that have up to $200$ total state features (\cf table \ref{tab:repartitions}). 
For these environments, the oblique tree extraction gives Out-of-Memory errors (reported as a random scores) when fitting oblique decision trees programs, during the main loop of \interpreter (line~\ref{alg:cart2} of algorithm~\ref{alg:iqdagger}). 
However, many consider features can be constant. For example, the $x$-coordinates of the player and the enemy in Pong, or the ladders coordinates in Kangaroo. 
We mask such features, as are incorporated in the decision boundaries of if-else conditions.
As shown in Table~\ref{tab:repartitions}, for Kangaroo and Seaquest, the number of feature can be divided (up to 5 times).
As depicted in Figure~\ref{fig:ablation}, there is a substantial performance gain obtained by masking idle features for \interpreter oblique tree programs' extraction in Atari environments.

\begin{figure}[t]
    \includegraphics[width=1\textwidth]{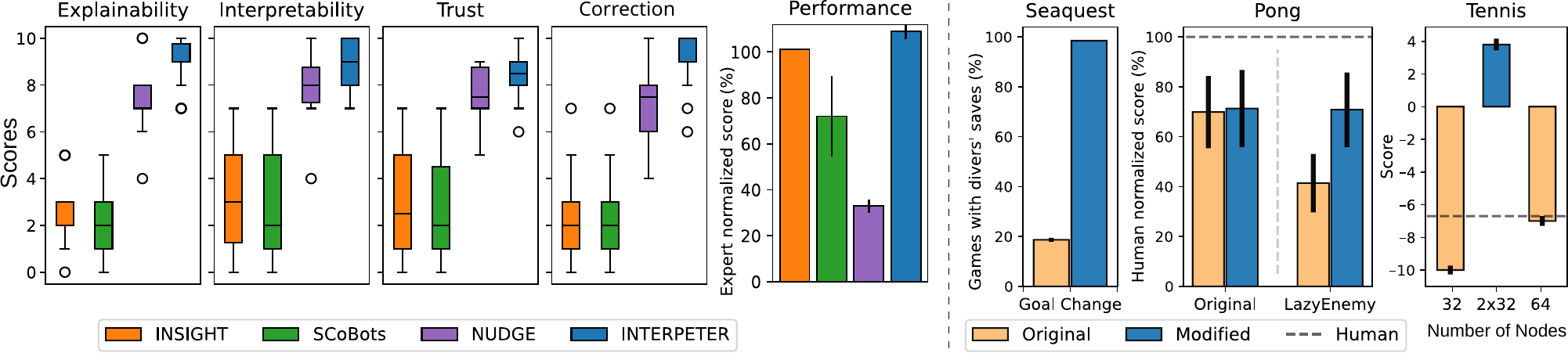}
    \caption{\textbf{\interpreter is the most interpretable policy form and can easily be modified.} Left: a user study shows that \interpreter is more explainable, interpretable, trustworthy and adjustable than other baselines, without sacrificing performances, contrary to the runner-up for these metrics (NUDGE). Right: Its Python policies allow for easy corrections on $3$ tasks variations.
    }
    \label{fig:experts}
\end{figure}

\subsection{\interpreter tree programs interpretability (Q3)}
\textbf{Inference speed.} One proxy for the interpretability of machine learning models without relying on human feedback is the inference computational complexity, which can be evaluated with code runtimes \citep{interpret_Lipton16a,barcelo2020model}.  
\interpreter extracts oblique decision trees and convert them to Python programs. 
We here compare the average inference speed of different policies during MDP rollouts, \ie how fast the policy outputs actions given states. 
All the inference speeds are measured without using pre-compiled code, neural network based oracles are instances of \texttt{PyTorch} modules~\citep{NEURIPS2019_bdbca288}. The \interpreter trees and programs have $64$ oblique test nodes. Results show that Python programs inference takes in average $79\mu s$, compared to $236\mu s$ for \texttt{scikit-learn} tree class and $466\mu s$ \texttt{PyTorch} networks (\cf Table~\ref{tab:inference-time} in the appendix for detailed results). 
The inference time is above all important for real world deployments of algorithms. As explain in the introduction, these deployments also require high interpretability and trust levels.

\textbf{User study.} We compared the interpretability and performances of \interpreter against {\small INSIGHT}~\citep{Luo2024INSIGHTEN}, SCoBots~\citep{delfosse2024interpretable} and {\small NUDGE}~\citep{Delfosse2023InterpretableAE}. We conducted a user study involving $19$ machine learning practitioners. They were asked to evaluate the \textit{explainability} (\ie ability to detect each input feature importance), \textit{interpretability} (\ie how each element is used by the policy to
select an action) and \textit{trust} (\ie ability to check if the agent selects the correct action for the correct reasons) levels of policies extracted by each method on Pong (as these policies are accessible for each method). 
Results are reported in Figure~\ref{fig:experts}, and show that \interpreter achieves the highest scores on each measurement. Contrary to its runner-up on these metrics ({\small NUDGE}), \interpreter does not sacrifice performances for interpretability.

\subsection{\interpreter tree programs edition (Q3)}
We further showcase $3$ modification possibilities on policies of \interpreter.

\textbf{Seaquest} contains ill-defined reward~\citep{delfosse2024interpretable}, as the game goal in the instruction manual is to ``retrieve as many treasure-divers as you can''\footnote{\href{https://atariage.com/manual_html_page.php?SoftwareLabelID=424}{https://atariage.com/instruction\_seaquest}. }. 
However, the game does provide any reward to the agent for saving each collected divers, but rather for killing enemies. 
Thus, both the neural and Python policies do not bring collected divers back to the surface. By simply adding:
\mintinline{Python}{if diver_0.x > 0: return "UP"}
\quad at the start of the 32 nodes \interpreter tree program, we correct this suboptimal behavior. 
As shown in Figure~\ref{fig:experts} (Left), the agents are able to save at least one diver in $98.5\%$ of the cases (compared to $18.6\%$ for the original agent). The complete and modified \interpreter program is provided in Figure~\ref{fig:seaquest-edited-code} in the Appendix.

\textbf{Pong.} RL policies can learn to achieve a misaligned goal, \ie to rely on the enemy's vertical position for their decision process (instead of the ball's one) \citep{delfosse2024interpretable}, as the two object's vertical positions are $99.9\%$ correlated. 
Our simple Pong program policy indeed uses \texttt{Enemy.y} in $1$ out of $6$ conditional tests (\cf the Pong policy in Figure~\ref{fig:pong-code}). 
They introduce a \textit{LazyEnemy} modified version of the environment, where the Enemy remains still after returning the ball, showing that many policies fail to generalize to this environment. We modify our policy by simply replacing \texttt{Enemy.y} with \texttt{Ball.y}, and test this modified policy on both environments. 
This leads to equivalent performances on the original training environment and prevents performance drops in the \textit{LazyEnemy} variation (\cf~Figure~\ref{fig:experts}). Compared to~\cite{delfosse2024interpretable}, that retrain an oracle, while hiding the enemy, we make one simple modification of the program, and do not need retraining.

\textbf{Tennis.} For this environment, {\small INTERPRETER}'s tree programs (\ie $16$ to $64$ nodes) cannot match the oracle's performances. 
Tennis is a complicated variation of Pong, as it adds the $y$-coordinates. However, oracle strategies can easily be divided into two sub-strategies, one where the agent is above the net (on the upper part of the screen), and one where it is under (on the lower side). 
To show the easy curriculum learning possibility offered by \interpreter, we extract two $32$ nodes-based code, based on a partition of the environment, based on the player's position. We are thus able to extract two policies, correctly performing only on one environment variation each, that we call within a meta policy. This policy calls each expert policy, depending on the evaluation of \mintinline{Python}{player.y - enemy.y > 0}. This meta-policy ($2$x$32$) outperforms the $32$ and $64$ nodes ones (\cf Figure~\ref{fig:experts}). We have shown that the \interpreter code-based policies allow for easy corrections.

\subsection{Real life use case of \interpreter tree programs for fertilization of soils (Q3)}
\begin{wrapfigure}[18]{r}{0.53\textwidth}
\centering
\vspace{-5.6mm}
\footnotesize
\begin{minted}{Python}
if nitrogen - days_planting < -17.50:
   if nitrogen - grain_weight < 13.50:
      if nitrogen - days_planting < -39.50:
         if nitrogen - maize_growing < -5.00:
            return "fertilizer_quant": 0.0
         else:
            
            return "fertilizer_quant": 27.0
      else:
         return "fertilizer_quant": 0.0
   else:
      if nitrogen - biomass < -930.64:
         return "fertilizer_quant": 54.0
      else:
         return "fertilizer_quant": 35.0
\end{minted}
\caption{\textbf{\interpreter can explain human heuristic policies.} \interpreter program with 100\% accuracy on human oracle policy.}
\vspace{-2mm}
\label{fig:dssat-iqdagger}
\end{wrapfigure}
In this last experiment, we distill a human expert policy for soil fertilization on the \texttt{gym-DSSAT gym} environment \cite{gautron2023learning}. Here, an RL agent has to learn to manage a crop, based on an accurate simulated mechanistic model of plant growth. 
We consider the task that consists in optimizing plant nitrogen absorption while penalizing the application of fertilizer to minimize the economical and the environmental costs. 
We extract an \interpreter's Python program, depicted in Figure~\ref{fig:dssat-iqdagger}. 
This program outputs the exact same actions as the human heuristic given the soil state and obtain the same cumulative reward in average (corresponding to an accuracy of $100\%$). 
It also provides an interpretation of the human expert heuristic that delivers a certain amount of fertilizer ($\{27, 35, 54\}$) after $\{39, 45, 80\}$ days after seeding, respectively). 
The feature importance coincides with agronomic principles and have been validated by an expert from the \textit{Consultative Group on International Agricultural Research}. 
The nitrogen requirements of corn vary depending on the growth stage. They are important during the vegetative phase (plant growth) and the reproductive phase (from flowering to grain filling). This is why it is essential to consider the number of days after planting and the growth stage of the corn, as nitrogen requirements are highest during grain filling.

\section{Related work}
\textbf{Explainable policies.} Algorithms for policy verification, such as the fast oracle matching VIPER~\citep{Bastani2018VerifiableRL}, only concentrate on optimizing an axes-parallel tree structure disregarding interpretability by growing many nodes trees. 
Thus, VIPER trees are explainable in the sense that one can always compute the set of rules verified by an MDP state that lead to an action, but deep trees are not interpretable in the simulatability sense 
\citep{interpret_Lipton16a}, because humans cannot themselves make the computations to explain actions chosen by the tree policy. 
Beyond VIPER, work from the neuro-symbolic RL community can learn explainable policies without oracles but require either carefully designed low level policy primitives to facilitate learning \citep{qiu2022programmatic} or large language models to attempt to explain learned policies \citep{Luo2024INSIGHTEN}.

\textbf{Interpretable policies.} On the other hand, some algorithms are designed to direcly optimize (with or without oracle knowledge) policies that are intrinsically interpretable. The recent SCoBots \citep{delfosse2024interpretable} output interpretable policies as sets of rules using {\small ECLAIRE}~\citep{zarlenga2021efficient} to match a PPO oracle. 
However, SCoBots require LLMs to define and experts to restrict the search space of sets of rules. 
Furthermore, LLMs might rely on external integrated knowledge about the game, acquired during their training phase, to explain the policy instead of providing accurate descriptions of it. 
Despite that, SCoBots, to the best of our knowledge, is the first algorithm that can consistently and automatically output interpretable policies for Atari games by using object-centric representations \citep{Bellemare2012TheAL,Delfosse2023OCAtariOA}. 
Prior to that, PIRL, Custard and NUDGE \citep{VermaMSKC18,IBMDP,Delfosse2023InterpretableAE} were also able to learn interpretable policies (programs, axes-parallel trees, and first-order logic respectively), but only on toy problems, for which environments were specifically created. 
Outside of RL, program synthesis has also been explored, \eg on classification tasks~\citep{ellis2021dreamcoder, Wst2024Pix2CodeLT}.

In addition to distinguishing existing work by the level of interpretabiliy of their returned policies and by their requirements for human and LLM interventions, we also distinguish the problems they can solve. {\small VIPER}, {\small NUDGE}, {\small INSIGHT} and Custard only work for MDPs with discrete actions, while the other algorithms work with any MDP. The above classification of existing interpretable RL algorithms are summarized in Appendix table \ref{tab:related-work}.
\newpage
\textbf{Symbolic states.} Recent interpretable RL assume access to a deep learning based object extractor that extract object-centric states from RGB inputs in game domains, such as SPACE~\citep{lin2020space}, SPOC~\citep{delfosse2021moc} or a finetuned FastSAM~\citep{Luo2024INSIGHTEN}. 
They train a neural network based policy using existing deep RL algorithm from this object-centric states. 
Then, they distill this policy into either directly interpretable (or transparent) first order logic-based NUDGE agents~\citep{Delfosse2023InterpretableAE} or rule-based SCoBots~\citep{delfosse2024interpretable}, or into (not interpretable) polynomial equations, within {\small INSIGHT}~\citep{Luo2024INSIGHTEN}.

\section{Limitations and future work}
\textbf{Decision tree algorithm.} 
CART \citep{breiman} is a widely used decision tree learning algorithm, however it chooses splits greedily w.r.t.\@ a train set, which is suboptimal \citep{Murthy}. There is a whole line of work on decision tree learning algorithm, some specialize in oblique trees \citep{murthy1994system}, others have better generalization performances \citep{quantbnb,murtree} and even better interpretability \citep{kohler2024interpretable}. One direct future direction for \interpreter would be to try different decision tree algorithms in algorithm \ref{alg:iqdagger}.

\textbf{More expressive and general tree programs.} Our \interpreter algorithm uses CART with linear combinations of at most two features as input, we could also add linear combinations of more features with coefficients different from $1$ or include more complex functions of features such as sinusoidal functions. 
It should also be possible to add an evolutionary routine \citep{eiben2015introduction} to include loops in the policy search space. One could for example try to gather rules applied on the same object types, to obtain \eg conditional tests on all the enemies or on every missile in environments like Seaquest or SpaceInvaders.

\textbf{Complexity and state space.} Exploring the solution space of policies defined over symbolic states is necessary for interpretabiliy but comes with limitations. For example, in Seaquest, if the oxygen bar level is not encoded in the symbolic states, the agent cannot learn to make decisions based on the latter. In MsPacman, the walls can be considered part of the background~\citep{lin2020space, delfosse2021moc}, but are necessary to navigate the maze. Generally, identifying what symbolic features are necessary to master each task is a difficult problem \citep{delfosse2024interpretable}. Furthermore, the complexity of the input space grows with the number of symbols in the state-space. We have proposed to mask idle state features, but there is no guarantee that this will sufficiently reduce the complexity for the decision tree extraction to be done in a limited time. One way to overcome this limitation would be to use \eg a deep learning alternatives that would return the oblique tests to consider for the tree programs given the whole state-action dataset \citep{kossen2021selfattention}, but such deep learning architecture complexity do not scale as much with the number of symbolic features. 

%Our algorithms present several avenues for future work. First, the use of only symbolic space is restrictive, as it can sometimes hide the complexity of the world. For example, in some environments, such as \textit{Pong}, walls can be ignored by the object extraction method, but they are crucial to the understanding and the solving of other tasks, such as \textit{MsPacman}. 
%Furthermore, while \interpreter produces editable tree programs, in which more complicated program-specific functionalities can be used (\eg loops, functions), it cannot natively produce general programs that include such forms. Future work might include the use of \eg LLM or logic algorithm to detect redundant parts of our tree programs and incorporate them into such structures. For example, it could detect that similar subtrees can be applied to object of the same types \eg divers or enemies in \textit{Seaquest}.

\textbf{Evaluating interpretability.} We have here evaluated the interpretability of \interpreter programs with an inference speed proxy and with limited user study. 
We should benchmark the interpretability of our programs with a more diverse pool of users, such as non machine learning practitioners, and could also make use of recently developed LLM code generation capabilities to explain our programs in natural language \citep{bashir2024assessing}. We then should evaluate the reliability of using LLMs to accurately explain the policy without relying on accumulated knowledge. 

\textbf{Code.} We have only provided the code to reproduce our experiments and a tutorial on a simple demo task. 
We are building a Python package to use \interpreter with \texttt{gym} and \texttt{PyTorch} oracles.

\section{Conclusion}
We have introduced \interpreter that distills deep RL oracle into interpretable programs to increase alignment and trust in automated sequential decision-making tasks. 
To do so, \interpreter produces tree programs that make oblique tests of meaningful state features. 
We empirically showed a state-of-the-art interpretability-performance trade-off: our programs match oracles and can be explained and edited by humans.
Furthermore, \interpreter is a simple algorithm: its components such as decision tree learning, features combinations, and programming language of the extracted programs, can be varied easily. 
For future work, we believe that benchmarking and safeguarding interpretability and alignment of policies to \eg human values are interesting avenues. We hope our work paves the way for future interpretable RL research.
\newpage
\bibliographystyle{plainnat}
\bibliography{tmlr}

\begin{thebibliography}{59}
\providecommand{\natexlab}[1]{#1}
\providecommand{\url}[1]{\texttt{#1}}
\expandafter\ifx\csname urlstyle\endcsname\relax
  \providecommand{\doi}[1]{doi: #1}\else
  \providecommand{\doi}{doi: \begingroup \urlstyle{rm}\Url}\fi

\bibitem[Arr(2020)]{Arrieta}
Explainable artificial intelligence (xai): Concepts, taxonomies, opportunities and challenges toward responsible ai.
\newblock \emph{Information Fusion}, 2020.

\bibitem[Barcel\'{o} et~al.(2020)Barcel\'{o}, Monet, P\'{e}rez, and Subercaseaux]{Barcelo20}
Pablo Barcel\'{o}, Mika\"{e}l Monet, Jorge P\'{e}rez, and Bernardo Subercaseaux.
\newblock Model interpretability through the lens of computational complexity.
\newblock In \emph{Neural Information Processing Systems (NeurIPS)}, 2020.

\bibitem[Barcel{\'o} et~al.(2020)Barcel{\'o}, Monet, P{\'e}rez, and Subercaseaux]{barcelo2020model}
Pablo Barcel{\'o}, Mika{\"e}l Monet, Jorge P{\'e}rez, and Bernardo Subercaseaux.
\newblock Model interpretability through the lens of computational complexity.
\newblock \emph{Advances in neural information processing systems}, 2020.

\bibitem[Bashir et~al.(2024)Bashir, Bowling, and Lelis]{bashir2024assessing}
Zahra Bashir, Michael Bowling, and Levi H.~S. Lelis.
\newblock Assessing the interpretability of programmatic policies with large language models, 2024.

\bibitem[Bastani et~al.(2018)Bastani, Pu, and Solar-Lezama]{Bastani2018VerifiableRL}
Osbert Bastani, Yewen Pu, and Armando Solar-Lezama.
\newblock Verifiable reinforcement learning via policy extraction.
\newblock In \emph{Neural Information Processing Systems}, 2018.

\bibitem[Bellemare et~al.(2012)Bellemare, Naddaf, Veness, and Bowling]{Bellemare2012TheAL}
Marc~G. Bellemare, Yavar Naddaf, Joel Veness, and Michael Bowling.
\newblock The arcade learning environment: An evaluation platform for general agents (extended abstract).
\newblock In \emph{International Joint Conference on Artificial Intelligence}, 2012.

\bibitem[Bhatt et~al.(2024)Bhatt, Palenicek, Belousov, Argus, Amiranashvili, Brox, and Peters]{bhatt2024crossq}
Aditya Bhatt, Daniel Palenicek, Boris Belousov, Max Argus, Artemij Amiranashvili, Thomas Brox, and Jan Peters.
\newblock Crossq: Batch normalization in deep reinforcement learning for greater sample efficiency and simplicity.
\newblock In \emph{The Twelfth International Conference on Learning Representations}, 2024.
\newblock URL \url{https://openreview.net/forum?id=PczQtTsTIX}.

\bibitem[Breiman et~al.(1984)Breiman, Friedman, Olshen, and Stone]{breiman}
Leo Breiman, Jerome Friedman, R.A. Olshen, and Charles~J. Stone.
\newblock \emph{Classification And Regression Trees}.
\newblock Taylor and Francis, New York, 1984.

\bibitem[Delfosse et~al.(2023{\natexlab{a}})Delfosse, Bl{\"u}ml, Gregori, Sztwiertnia, and Kersting]{Delfosse2023OCAtariOA}
Quentin Delfosse, Jannis Bl{\"u}ml, Bjarne Gregori, Sebastian Sztwiertnia, and Kristian Kersting.
\newblock Ocatari: Object-centric atari 2600 reinforcement learning environments.
\newblock \emph{ArXiv}, 2023{\natexlab{a}}.

\bibitem[Delfosse et~al.(2023{\natexlab{b}})Delfosse, Shindo, Dhami, and Kersting]{Delfosse2023InterpretableAE}
Quentin Delfosse, Hikaru Shindo, Devendra~Singh Dhami, and Kristian Kersting.
\newblock Interpretable and explainable logical policies via neurally guided symbolic abstraction.
\newblock 2023{\natexlab{b}}.

\bibitem[Delfosse et~al.(2023{\natexlab{c}})Delfosse, Stammer, Rothenbacher, Vittal, and Kersting]{delfosse2021moc}
Quentin Delfosse, Wolfgang Stammer, Thomas Rothenbacher, Dwarak Vittal, and Kristian Kersting.
\newblock Boosting object representation learning via motion and object continuity.
\newblock In Danai Koutra, Claudia Plant, Manuel~Gomez Rodriguez, Elena Baralis, and Francesco Bonchi, editors, \emph{European Conference on Machine Learning and Principles and Practice of Knowledge Discovery in Databases ({ECML})}, 2023{\natexlab{c}}.

\bibitem[Delfosse et~al.(2024)Delfosse, Sztwiertnia, Stammer, Rothermel, and Kersting]{delfosse2024interpretable}
Quentin Delfosse, Sebastian Sztwiertnia, Wolfgang Stammer, Mark Rothermel, and Kristian Kersting.
\newblock Interpretable concept bottlenecks to align reinforcement learning agents.
\newblock \emph{arXiv}, 2024.

\bibitem[Demirovic et~al.(2022)Demirovic, Lukina, Hebrard, Chan, Bailey, Leckie, Ramamohanarao, and Stuckey]{murtree}
Emir Demirovic, Anna Lukina, Emmanuel Hebrard, Jeffrey Chan, James Bailey, Christopher Leckie, Kotagiri Ramamohanarao, and Peter~J. Stuckey.
\newblock Murtree: Optimal decision trees via dynamic programming and search.
\newblock \emph{Journal of Machine Learning Research}, 23\penalty0 (26):\penalty0 1--47, 2022.
\newblock URL \url{http://jmlr.org/papers/v23/20-520.html}.

\bibitem[di~Langosco et~al.(2022)di~Langosco, Koch, Sharkey, Pfau, and Krueger]{Langosco2022goal}
Lauro~Langosco di~Langosco, Jack Koch, Lee~D. Sharkey, Jacob Pfau, and David Krueger.
\newblock Goal misgeneralization in deep reinforcement learning.
\newblock In \emph{International Conference on Machine Learning {ICML}}, 2022.

\bibitem[Eiben and Smith(2015)]{eiben2015introduction}
Agoston~E Eiben and James~E Smith.
\newblock \emph{Introduction to evolutionary computing}.
\newblock Springer, 2015.

\bibitem[Ellis et~al.(2021)Ellis, Wong, Nye, Sabl{\'e}-Meyer, Morales, Hewitt, Cary, Solar-Lezama, and Tenenbaum]{ellis2021dreamcoder}
Kevin Ellis, Catherine Wong, Maxwell Nye, Mathias Sabl{\'e}-Meyer, Lucas Morales, Luke Hewitt, Luc Cary, Armando Solar-Lezama, and Joshua~B Tenenbaum.
\newblock Dreamcoder: Bootstrapping inductive program synthesis with wake-sleep library learning.
\newblock In \emph{Proceedings of the 42nd acm sigplan international conference on programming language design and implementation}, 2021.

\bibitem[Gautron et~al.(2023)Gautron, Padr{\'o}n, Preux, Bigot, Maillard, Hoogenboom, and Teigny]{gautron2023learning}
Romain Gautron, Emilio~J Padr{\'o}n, Philippe Preux, Julien Bigot, Odalric-Ambrym Maillard, Gerrit Hoogenboom, and Julien Teigny.
\newblock Learning crop management by reinforcement: gym-dssat.
\newblock In \emph{AIAFS 2023-2nd AAAI Workshop on AI for Agriculture and Food Systems}, 2023.

\bibitem[Glanois et~al.(2021)Glanois, Weng, Zimmer, Li, Yang, Hao, and Liu]{glanois2022xrl}
Claire Glanois, P.~Weng, Matthieu Zimmer, Dong Li, Tianpei Yang, Jianye Hao, and Wulong Liu.
\newblock A survey on interpretable reinforcement learning.
\newblock \emph{ArXiv}, 2021.

\bibitem[Guidotti et~al.(2018)Guidotti, Monreale, Ruggieri, Turini, Giannotti, and Pedreschi]{Guidotti18}
Riccardo Guidotti, Anna Monreale, Salvatore Ruggieri, Franco Turini, Fosca Giannotti, and Dino Pedreschi.
\newblock A survey of methods for explaining black box models.
\newblock \emph{ACM Comput. Surv.}, 2018.

\bibitem[Haarnoja et~al.(2018)Haarnoja, Zhou, Abbeel, and Levine]{pmlr-v80-haarnoja18b}
Tuomas Haarnoja, Aurick Zhou, Pieter Abbeel, and Sergey Levine.
\newblock Soft actor-critic: Off-policy maximum entropy deep reinforcement learning with a stochastic actor.
\newblock In \emph{Proceedings of the 35th International Conference on Machine Learning}, 2018.

\bibitem[Kohler et~al.(2024)Kohler, Akrour, and Preux]{kohler2024interpretable}
Hector Kohler, Riad Akrour, and Philippe Preux.
\newblock Interpretable decision tree search as a markov decision process, 2024.

\bibitem[Kossen et~al.(2021)Kossen, Band, Lyle, Gomez, Rainforth, and Gal]{kossen2021selfattention}
Jannik Kossen, Neil Band, Clare Lyle, Aidan Gomez, Tom Rainforth, and Yarin Gal.
\newblock Self-attention between datapoints: Going beyond individual input-output pairs in deep learning.
\newblock In A.~Beygelzimer, Y.~Dauphin, P.~Liang, and J.~Wortman Vaughan, editors, \emph{Advances in Neural Information Processing Systems}, 2021.
\newblock URL \url{https://openreview.net/forum?id=wRXzOa2z5T}.

\bibitem[Krizhevsky et~al.(2012)Krizhevsky, Sutskever, and Hinton]{AlexNet}
Alex Krizhevsky, Ilya Sutskever, and Geoffrey~E Hinton.
\newblock Imagenet classification with deep convolutional neural networks.
\newblock In F.~Pereira, C.J. Burges, L.~Bottou, and K.Q. Weinberger, editors, \emph{Advances in Neural Information Processing Systems}, 2012.

\bibitem[Landajuela et~al.(2021)Landajuela, Petersen, Kim, Santiago, Glatt, Mundhenk, Pettit, and Faissol]{landajuela}
Mikel Landajuela, Brenden~K Petersen, Sookyung Kim, Claudio~P Santiago, Ruben Glatt, Nathan Mundhenk, Jacob~F Pettit, and Daniel Faissol.
\newblock Discovering symbolic policies with deep reinforcement learning.
\newblock In \emph{International Conference on Machine Learning}, pages 5979--5989. PMLR, 2021.

\bibitem[Lin et~al.(2020)Lin, Wu, Peri, Sun, Singh, Deng, Jiang, and Ahn]{lin2020space}
Zhixuan Lin, Yi{-}Fu Wu, Skand~Vishwanath Peri, Weihao Sun, Gautam Singh, Fei Deng, Jindong Jiang, and Sungjin Ahn.
\newblock {SPACE:} unsupervised object-oriented scene representation via spatial attention and decomposition.
\newblock In \emph{International Conference on Learning Representations}, 2020.

\bibitem[Lipton(2016)]{interpret_Lipton16a}
Zachary~Chase Lipton.
\newblock The mythos of model interpretability.
\newblock \emph{ArXiv}, 2016.

\bibitem[Luo et~al.(2024)Luo, Zhang, Xu, Yang, Fang, and Li]{Luo2024INSIGHTEN}
Lirui Luo, Guoxi Zhang, Hongming Xu, Yaodong Yang, Cong Fang, and Qing Li.
\newblock Insight: End-to-end neuro-symbolic visual reinforcement learning with language explanations.
\newblock \emph{ArXiv}, 2024.

\bibitem[Machado et~al.(2018)Machado, Bellemare, Talvitie, Veness, Hausknecht, and Bowling]{MachadoBTVHB18}
Marlos~C. Machado, Marc~G. Bellemare, Erik Talvitie, Joel Veness, Matthew~J. Hausknecht, and Michael Bowling.
\newblock Revisiting the arcade learning environment: Evaluation protocols and open problems for general agents (extended abstract).
\newblock In \emph{Proceedings of the Twenty-Seventh International Joint Conference on Artificial Intelligence, {IJCAI}}, 2018.

\bibitem[Mazumder et~al.(2022)Mazumder, Meng, and Wang]{quantbnb}
Rahul Mazumder, Xiang Meng, and Haoyue Wang.
\newblock Quant-{B}n{B}: A scalable branch-and-bound method for optimal decision trees with continuous features.
\newblock In Kamalika Chaudhuri, Stefanie Jegelka, Le~Song, Csaba Szepesvari, Gang Niu, and Sivan Sabato, editors, \emph{Proceedings of the 39th International Conference on Machine Learning}, volume 162 of \emph{Proceedings of Machine Learning Research}, pages 15255--15277. PMLR, 17--23 Jul 2022.
\newblock URL \url{https://proceedings.mlr.press/v162/mazumder22a.html}.

\bibitem[Milani et~al.(2022)Milani, Topin, Veloso, and Fang]{milani2022survey}
Stephanie Milani, Nicholay Topin, Manuela~M. Veloso, and Fei Fang.
\newblock A survey of explainable reinforcement learning.
\newblock \emph{ArXiv}, 2022.

\bibitem[Mnih et~al.(2013)Mnih, Kavukcuoglu, Silver, Graves, Antonoglou, Wierstra, and Riedmiller]{Mnih2013PlayingAW}
Volodymyr Mnih, Koray Kavukcuoglu, David Silver, Alex Graves, Ioannis Antonoglou, Daan Wierstra, and Martin~A. Riedmiller.
\newblock Playing atari with deep reinforcement learning.
\newblock \emph{ArXiv}, 2013.

\bibitem[Mnih et~al.(2015)Mnih, Kavukcuoglu, Silver, Rusu, Veness, Bellemare, Graves, Riedmiller, Fidjeland, Ostrovski, Petersen, Beattie, Sadik, Antonoglou, King, Kumaran, Wierstra, Legg, and Hassabis]{Mnih2015dqn}
Volodymyr Mnih, Koray Kavukcuoglu, David Silver, Andrei~A. Rusu, Joel Veness, Marc~G. Bellemare, Alex Graves, Martin~A. Riedmiller, Andreas Fidjeland, Georg Ostrovski, Stig Petersen, Charles Beattie, Amir Sadik, Ioannis Antonoglou, Helen King, Dharshan Kumaran, Daan Wierstra, Shane Legg, and Demis Hassabis.
\newblock Human-level control through deep reinforcement learning.
\newblock \emph{Nature}, 2015.

\bibitem[Murdoch et~al.(2019)Murdoch, Singh, Kumbier, Abbasi-Asl, and Yu]{Murdoch}
W.~James Murdoch, Chandan Singh, Karl Kumbier, Reza Abbasi-Asl, and Bin Yu.
\newblock Definitions, methods, and applications in interpretable machine learning.
\newblock \emph{Proceedings of the National Academy of Sciences}, 2019.

\bibitem[Murthy and Salzberg(1995)]{Murthy}
Sreerama Murthy and Steven Salzberg.
\newblock Lookahead and pathology in decision tree induction.
\newblock In \emph{Proceedings of the 14th International Joint Conference on Artificial Intelligence - Volume 2}, IJCAI'95, page 1025–1031, San Francisco, CA, USA, 1995. Morgan Kaufmann Publishers Inc.
\newblock ISBN 1558603638.

\bibitem[Murthy et~al.(1994)Murthy, Kasif, and Salzberg]{murthy1994system}
Sreerama~K Murthy, Simon Kasif, and Steven Salzberg.
\newblock A system for induction of oblique decision trees.
\newblock \emph{Journal of artificial intelligence research}, 1994.

\bibitem[Paszke et~al.(2019)Paszke, Gross, Massa, Lerer, Bradbury, Chanan, Killeen, Lin, Gimelshein, Antiga, Desmaison, Kopf, Yang, DeVito, Raison, Tejani, Chilamkurthy, Steiner, Fang, Bai, and Chintala]{NEURIPS2019_bdbca288}
Adam Paszke, Sam Gross, Francisco Massa, Adam Lerer, James Bradbury, Gregory Chanan, Trevor Killeen, Zeming Lin, Natalia Gimelshein, Luca Antiga, Alban Desmaison, Andreas Kopf, Edward Yang, Zachary DeVito, Martin Raison, Alykhan Tejani, Sasank Chilamkurthy, Benoit Steiner, Lu~Fang, Junjie Bai, and Soumith Chintala.
\newblock Pytorch: An imperative style, high-performance deep learning library.
\newblock In \emph{Advances in Neural Information Processing Systems}. Curran Associates, Inc., 2019.

\bibitem[Pedregosa et~al.(2011)Pedregosa, Varoquaux, Gramfort, Michel, Thirion, Grisel, Blondel, Prettenhofer, Weiss, Dubourg, Vanderplas, Passos, Cournapeau, Brucher, Perrot, and Duchesnay]{scikit-learn}
F.~Pedregosa, G.~Varoquaux, A.~Gramfort, V.~Michel, B.~Thirion, O.~Grisel, M.~Blondel, P.~Prettenhofer, R.~Weiss, V.~Dubourg, J.~Vanderplas, A.~Passos, D.~Cournapeau, M.~Brucher, M.~Perrot, and E.~Duchesnay.
\newblock Scikit-learn: Machine learning in {P}ython.
\newblock \emph{Journal of Machine Learning Research}, 2011.

\bibitem[Puterman(2014)]{puterman2014markov}
Martin~L Puterman.
\newblock \emph{Markov decision processes: discrete stochastic dynamic programming}.
\newblock John Wiley \& Sons, 2014.

\bibitem[Qiu and Zhu(2022)]{qiu2022programmatic}
Wenjie Qiu and He~Zhu.
\newblock Programmatic reinforcement learning without oracles.
\newblock In \emph{International Conference on Learning Representations}, 2022.

\bibitem[Raffin(2020)]{rl-zoo3}
Antonin Raffin.
\newblock Rl baselines3 zoo, 2020.

\bibitem[Raffin et~al.(2021)Raffin, Hill, Gleave, Kanervisto, Ernestus, and Dormann]{stable-baselines3}
Antonin Raffin, Ashley Hill, Adam Gleave, Anssi Kanervisto, Maximilian Ernestus, and Noah Dormann.
\newblock Stable-baselines3: Reliable reinforcement learning implementations.
\newblock \emph{Journal of Machine Learning Research}, 2021.

\bibitem[Ras et~al.(2022)Ras, Xie, van Gerven, and Doran]{RasXGD22}
Gabrielle Ras, Ning Xie, Marcel van Gerven, and Derek Doran.
\newblock Explainable deep learning: {A} field guide for the uninitiated.
\newblock \emph{Journal of Artificial Intelligence Research}, 2022.

\bibitem[Ross et~al.(2010)Ross, Gordon, and Bagnell]{Ross2010ARO}
St{\'e}phane Ross, Geoffrey~J. Gordon, and J.~Andrew Bagnell.
\newblock A reduction of imitation learning and structured prediction to no-regret online learning.
\newblock In \emph{International Conference on Artificial Intelligence and Statistics}, 2010.

\bibitem[Roy et~al.(2022)Roy, Kim, and Rabinowitz]{RoyKR22}
Nicholas~A. Roy, Junkyung Kim, and Neil~C. Rabinowitz.
\newblock Explainability via causal self-talk.
\newblock 2022.

\bibitem[Saeed and Omlin(2023)]{SaeedO23}
Waddah Saeed and Christian~W. Omlin.
\newblock Explainable {AI} {(XAI):} {A} systematic meta-survey of current challenges and future opportunities.
\newblock \emph{Knowledge-Based Systems}, 2023.

\bibitem[Sani et~al.(2018)Sani, Lei, and Neagu]{complexcart}
Habiba Sani, Ci~Lei, and Daniel Neagu.
\newblock \emph{Computational Complexity Analysis of Decision Tree Algorithms: 38th SGAI International Conference on Artificial Intelligence, AI 2018, Cambridge, UK, December 11–13, 2018, Proceedings}, pages 191--197.
\newblock 11 2018.
\newblock ISBN 978-3-030-04190-8.
\newblock \doi{10.1007/978-3-030-04191-5_17}.

\bibitem[Schramowski et~al.(2020)Schramowski, Stammer, Teso, Brugger, Herbert, Shao, Luigs, Mahlein, and Kersting]{SchramowskiSTBH20}
Patrick Schramowski, Wolfgang Stammer, Stefano Teso, Anna Brugger, Franziska Herbert, Xiaoting Shao, Hans{-}Georg Luigs, Anne{-}Katrin Mahlein, and Kristian Kersting.
\newblock Making deep neural networks right for the right scientific reasons by interacting with their explanations.
\newblock \emph{Nature Machine Intelligence}, 2020.

\bibitem[Schulman et~al.(2015)Schulman, Levine, Abbeel, Jordan, and Moritz]{pmlr-v37-schulman15}
John Schulman, Sergey Levine, Pieter Abbeel, Michael Jordan, and Philipp Moritz.
\newblock Trust region policy optimization.
\newblock In \emph{Proceedings of the 32nd International Conference on Machine Learning}, 2015.

\bibitem[Schulman et~al.(2017)Schulman, Wolski, Dhariwal, Radford, and Klimov]{Schulman2017ProximalPO}
John Schulman, Filip Wolski, Prafulla Dhariwal, Alec Radford, and Oleg Klimov.
\newblock Proximal policy optimization algorithms.
\newblock \emph{ArXiv}, 2017.

\bibitem[Sutton and Barto(2018)]{sutton2018reinforcement}
Richard~S Sutton and Andrew~G Barto.
\newblock \emph{Reinforcement learning: An introduction}.
\newblock MIT press, 2018.

\bibitem[Todorov et~al.(2012)Todorov, Erez, and Tassa]{todorov12mujoco}
Emanuel Todorov, Tom Erez, and Yuval Tassa.
\newblock Mujoco: A physics engine for model-based control.
\newblock In \emph{International Conference on Intelligent Robots andSystems (IROS)}, pages 5026--5033. IEEE, 2012.

\bibitem[Topin et~al.(2021)Topin, Milani, Fang, and Veloso]{IBMDP}
Nicholay Topin, Stephanie Milani, Fei Fang, and Manuela Veloso.
\newblock Iterative bounding mdps: Learning interpretable policies via non-interpretable methods.
\newblock \emph{Proceedings of the AAAI Conference on Artificial Intelligence}, 2021.

\bibitem[Towers et~al.(2023)Towers, Terry, Kwiatkowski, Balis, Cola, Deleu, Goulão, Kallinteris, KG, Krimmel, Perez-Vicente, Pierré, Schulhoff, Tai, Shen, and Younis]{towers_gymnasium_2023}
Mark Towers, Jordan~K. Terry, Ariel Kwiatkowski, John~U. Balis, Gianluca~de Cola, Tristan Deleu, Manuel Goulão, Andreas Kallinteris, Arjun KG, Markus Krimmel, Rodrigo Perez-Vicente, Andrea Pierré, Sander Schulhoff, Jun~Jet Tai, Andrew Tan~Jin Shen, and Omar~G. Younis.
\newblock Gymnasium, 2023.

\bibitem[van Hasselt et~al.(2016)van Hasselt, Guez, and Silver]{vanHasseltGS16ddqn}
Hado van Hasselt, Arthur Guez, and David Silver.
\newblock Deep reinforcement learning with double q-learning.
\newblock In \emph{Proceedings of the Thirtieth {AAAI} Conference on Artificial Intelligence, February 12-17, 2016, Phoenix, Arizona, {USA}}, 2016.

\bibitem[Vaswani et~al.(2017)Vaswani, Shazeer, Parmar, Uszkoreit, Jones, Gomez, Kaiser, and Polosukhin]{attention}
Ashish Vaswani, Noam Shazeer, Niki Parmar, Jakob Uszkoreit, Llion Jones, Aidan~N Gomez, \L~ukasz Kaiser, and Illia Polosukhin.
\newblock Attention is all you need.
\newblock In I.~Guyon, U.~Von Luxburg, S.~Bengio, H.~Wallach, R.~Fergus, S.~Vishwanathan, and R.~Garnett, editors, \emph{Advances in Neural Information Processing Systems}, volume~30. Curran Associates, Inc., 2017.
\newblock URL \url{https://proceedings.neurips.cc/paper_files/paper/2017/file/3f5ee243547dee91fbd053c1c4a845aa-Paper.pdf}.

\bibitem[Verma et~al.(2018)Verma, Murali, Singh, Kohli, and Chaudhuri]{VermaMSKC18}
Abhinav Verma, Vijayaraghavan Murali, Rishabh Singh, Pushmeet Kohli, and Swarat Chaudhuri.
\newblock Programmatically interpretable reinforcement learning.
\newblock In \emph{Proceedings of the 35th International Conference on Machine Learning, {ICML}}, 2018.

\bibitem[W{\"u}st et~al.(2024)W{\"u}st, Stammer, Delfosse, Dhami, and Kersting]{Wst2024Pix2CodeLT}
Antonia W{\"u}st, Wolfgang Stammer, Quentin Delfosse, Devendra~Singh Dhami, and Kristian Kersting.
\newblock Pix2code: Learning to compose neural visual concepts as programs.
\newblock \emph{ArXiv}, 2024.

\bibitem[Zarlenga et~al.(2021)Zarlenga, Shams, and Jamnik]{zarlenga2021efficient}
Mateo~Espinosa Zarlenga, Zohreh Shams, and Mateja Jamnik.
\newblock Efficient decompositional rule extraction for deep neural networks.
\newblock In \emph{eXplainable AI approaches for debugging and diagnosis.}, 2021.

\bibitem[Zhao et~al.(2023)Zhao, Ding, An, Du, Yu, Li, Tang, and Wang]{zhao2023fast}
Xu~Zhao, Wenchao Ding, Yongqi An, Yinglong Du, Tao Yu, Min Li, Ming Tang, and Jinqiao Wang.
\newblock Fast segment anything.
\newblock \emph{arXiv}, 2023.

\end{thebibliography}

\newpage

\appendix
\section{Neural oracles}
\begin{figure}[h!]
     \centering
     \begin{subfigure}[b]{0.245\textwidth}
         \centering
         \includegraphics[width=\textwidth]{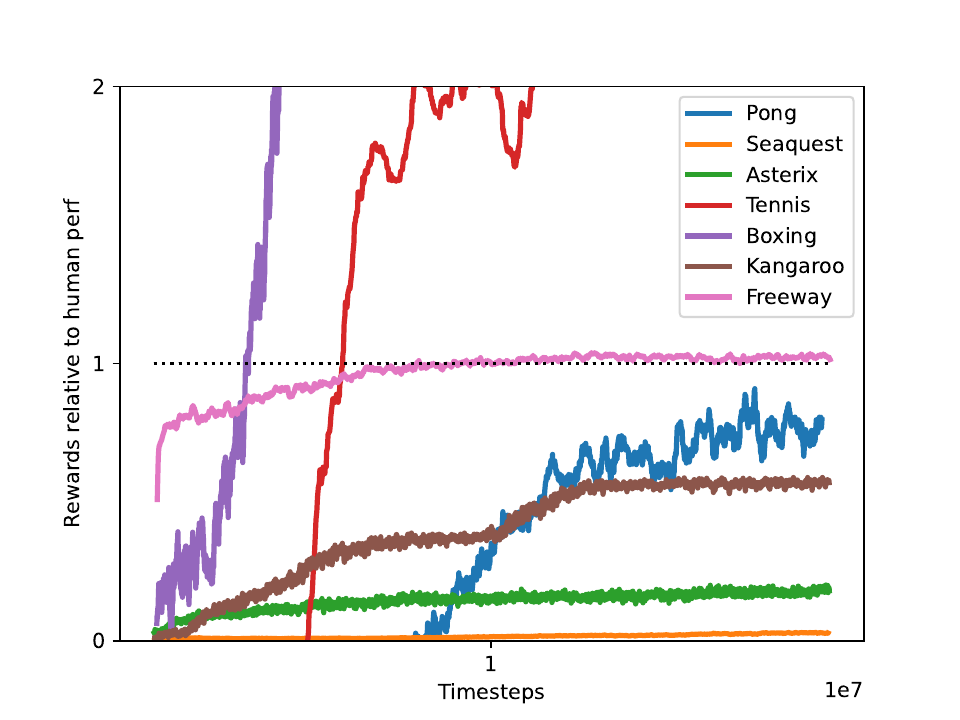}
     \end{subfigure}
     \begin{subfigure}[b]{0.245\textwidth}
         \centering
         \includegraphics[width=\textwidth]{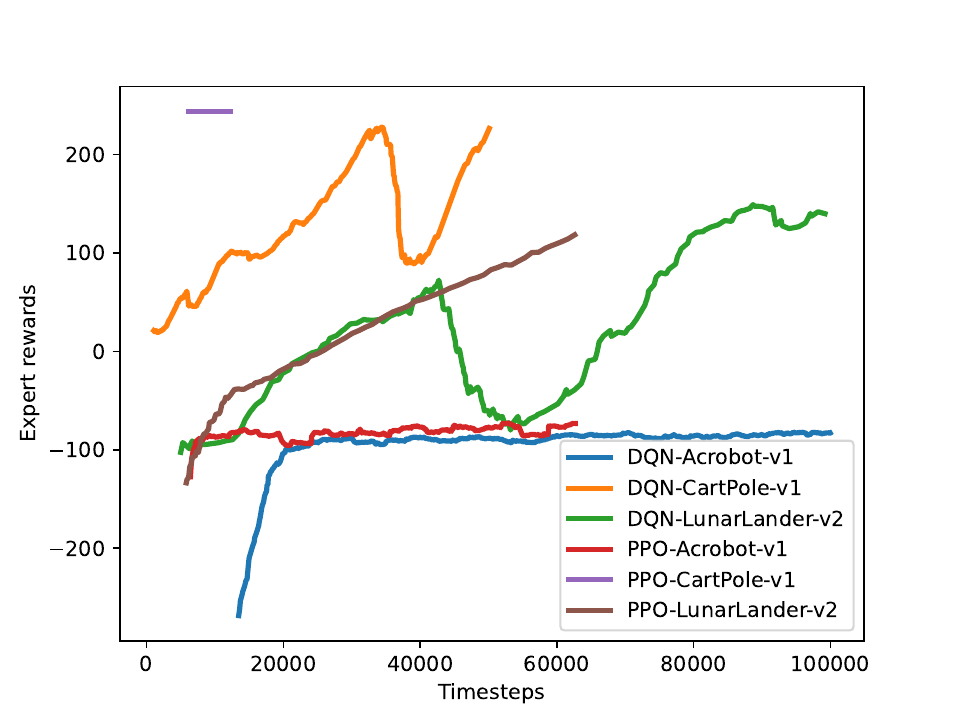}
     \end{subfigure}
     \begin{subfigure}[b]{0.245\textwidth}
         \centering
         \includegraphics[width=\textwidth]{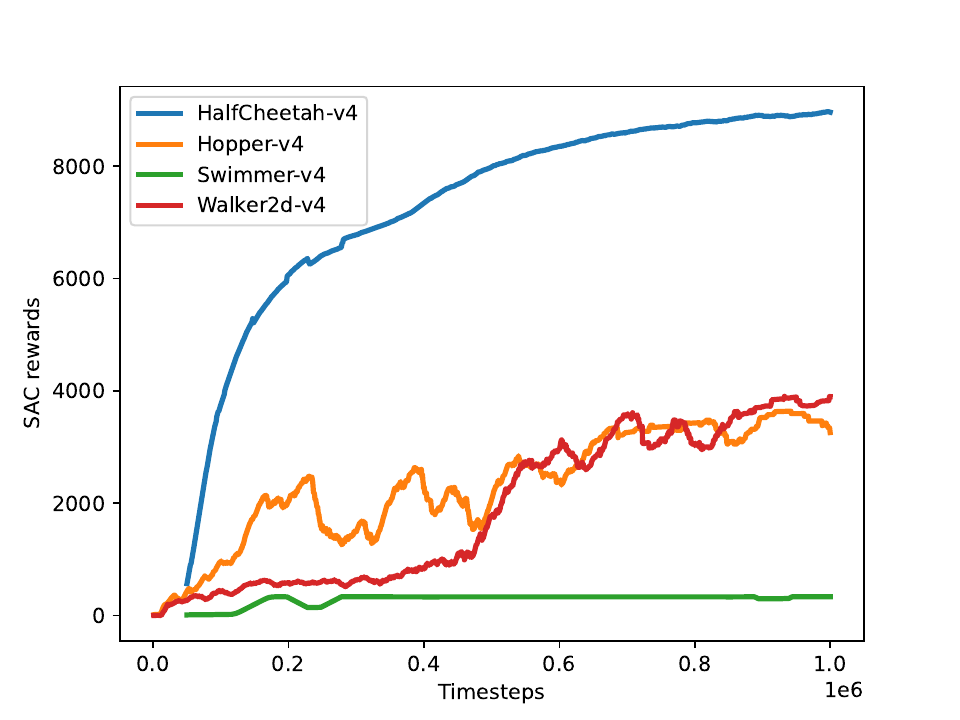}
     \end{subfigure}
     \begin{subfigure}[b]{0.245\textwidth}
         \centering
         \includegraphics[width=\textwidth]{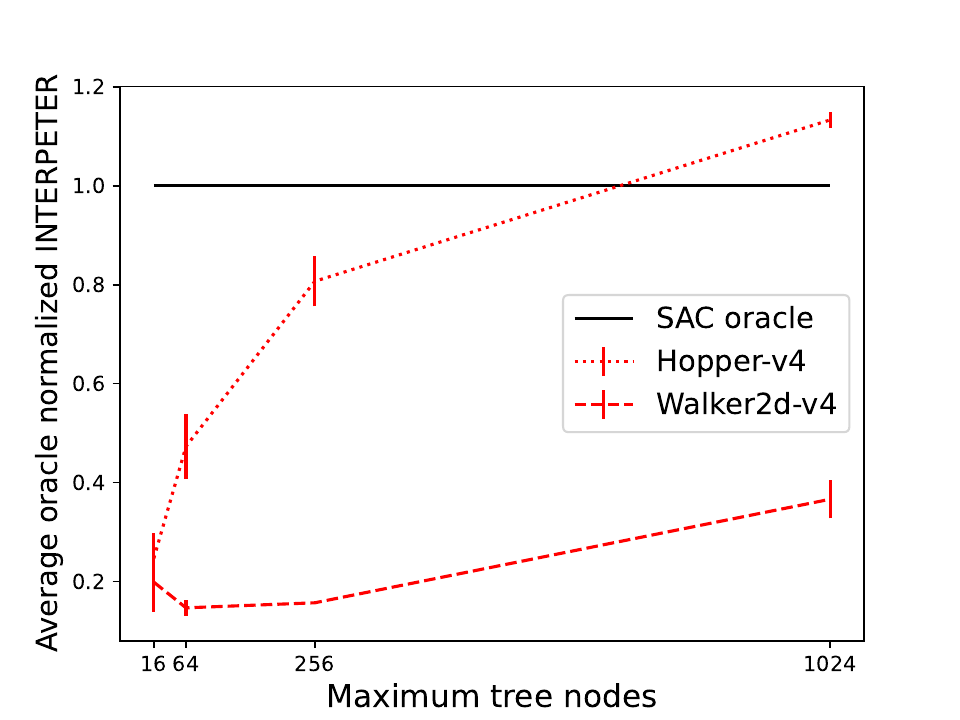}
     \end{subfigure}
     \caption{Detailed oracle training curves for Atari, mujoco and classic control environments, as well as the performance evolution of the oracles for different tree sizes for complex control tasks.}
    \label{fig:summary-oracles}
\end{figure}

In this section we show the training curves of the neural oracles used in sections \ref{sec:performances}, \ref{sec:algo-abl}. For the the MuJoCo and classic control benchmarks (center and right on Figure~\ref{fig:summary-oracles}), the neural oracles policies as well as training data are taken directly from the \texttt{stable-baselines3 zoo}, i.e we do not run the training ourselves. For example, all the data for the SAC oracle on Swimmer can be found at this url \url{https://github.com/DLR-RM/rl-trained-agents/tree/ca4371d8eef7c2eece81461f3d138d23743b2296/sac/Swimmer-v3_1}. For OC Atari we train the PPO agent from \texttt{stable-baselines3} ourselves on a DGX cluster. We use the default \citep{Schulman2017ProximalPO} hyperparameters on 2e7 timesteps. What we observe is that for most benchmarks the deep reinforcement learning algorithms converged except for the DQN agents on classic control tasks (Figure~\ref{fig:summary-oracles}, center). We also show that some SAC oracle are too complex to be matched by oblique tree prorgrams even high a high number of nodes (right of Figure~\ref{fig:summary-oracles}).

\section{Per game ablation}
\begin{figure}[h!]
% \vspace*{-1cm}  
    \includegraphics[width=1\textwidth]{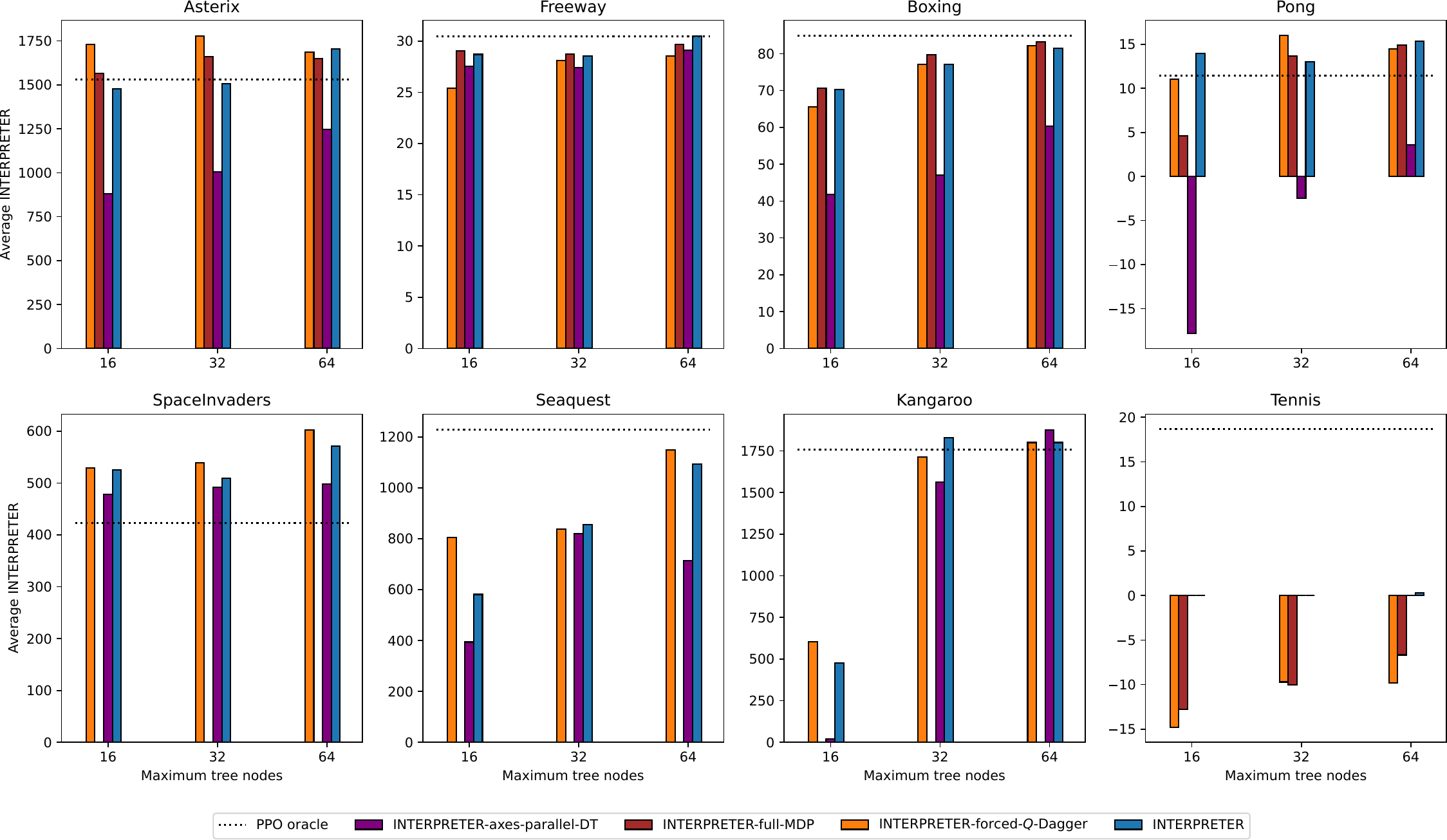}
    \caption{Detail ablation of \interpreter.}
    \label{fig:detailed-ablation}
\end{figure}
In this section, we ablate \interpreter on OC Atari games. We remove each element of method \ref{sec:algo} to get four distinct algorithms. Figure~\ref{fig:detailed-ablation} clearly shows that each independent component of \interpreter participates in its performances.

\textbf{\interpreter}: the default implementation of algorithm \ref{alg:iqdagger}.

\textbf{{\small INTERPRETER-axes-parallel-DT}}: we fit axes-parallel trees to oracles rather than oblique ones.

\textbf{{\small INTERPRETER-full-MDP}}: we do not mask idle MDP features during the imitation. In addition the performances we show in \ref{tab:repartitions} the share of idle features for each game.

\textbf{{\small INTERPRETER-forced-$Q$-Dagger}}: even though the given oracle is the stochastic policy $\pi^*$ returned by a PPO, we also pass $Q^{\pi^*}=\log\pi^*$ to "force" the $Q$-Dagger weighting during the imitation subroutine. 
\section{Detailed inference time per game}
\begin{table}[h!]
\centering
\caption{\textbf{INTERPRETER Programs are more resource efficient.} Inference speed comparisons (in $10^{-5} s$) of programs, \texttt{scikit-learn} decision trees and \texttt{PyTorch} netowrks on different gym environments. For all environments, our compact programs show faster inference time. }
\begin{tabular}{@{}lllllll@{}}
\toprule
                      & Asterix              & Boxing               & Freeway              & Kanga.             & Pong                 & Seaquest             \\ \midrule
Program               & 1.05\tiny$\pm$0.44   & 1.47\tiny$\pm$0.58  & 0.85\tiny$\pm$0.24  & 1.24\tiny$\pm$0.33  & 1.01 \tiny$\pm$0.24 & 0.97\tiny$\pm$0.17  \\
\texttt{sklearn} tree & 26.0\tiny$\pm$1.69 & 30.6\tiny$\pm$5.91 & 20.8\tiny$\pm$0.42 & 21.7\tiny$\pm$0.41 & 20.4\tiny$\pm$1.78 & 23.1\tiny$\pm$0.49 \\
Neural Net.           & 38.1\tiny$\pm$6.5   & 58.5\tiny$\pm$14.4  & 38.5\tiny$\pm$5.3   & 38.4\tiny$\pm$8.8   & 33.2 \tiny$\pm$6.5  & 38.3\tiny$\pm$7.25  \\ \bottomrule
\end{tabular}
\begin{tabular}{@{}lllllll@{}}
\toprule
                      & SpaceInv       & Tennis               & HalfCh.          & Hopper                & Swimmer               & Walker2d              \\ \midrule
Program               & 1.07\tiny$\pm$0.22  & 0.9\tiny$\pm$0.36 & 0.18\tiny$\pm$0.04  & 0.25\tiny$\pm$0.04   & 0.19\tiny$\pm$0.03   & 0.25\tiny$\pm$0.06   \\
\texttt{sklearn} tree & 40.9\tiny$\pm$9.22 & 20.3\tiny$\pm$0.45 & 16.5\tiny$\pm$1.72  & 20.2\tiny$\pm$1.5  & 19.4\tiny$\pm$1.3  & 23.8\tiny$\pm$4.01  \\
Neural Net.           & 41.9\tiny$\pm$10.4  & 34.8\tiny$\pm$2.8  & 54.2\tiny$\pm$24.0 & 66.6\tiny$\pm$22.2 & 56.3\tiny$\pm$20.8 & 60.9\tiny$\pm$18.5 \\ \bottomrule
\end{tabular}
\label{tab:inference-time}
\end{table}

\section{Feature importances for shortcut learning identification.}
% \begin{wrapfigure}{r}{0.3\textwidth}
%     \centering
%     \includegraphics[width=0.2\textwidth]{figures/hopper.jpeg}
%     \caption{Hopper simulation}
%     \label{fig:hopper}
% \end{wrapfigure}
In this section we look at the feature importances of tree programs with 16 total nodes returned by \interpreter on MuJoCo and Atari. 
Some clear importances are for the Hopper robot that needs to move forward by jumping on one leg: \interpreter bases its control on the $z$-coordinates of the torso. Some other clear importances are: for Pong where program bases its decision on the $y$-distance between the player's pad and the ball; for Freeway where the most important is the chicken's $y$-speed; for Asterix it is the $y$-distance to the collectible...
For Seaquest and Kangaroo where the goals are respectively to save divers and to get up to save its joey, we notice that the most important \interpreter concepts for the oracle do not include those goals. When visualizing the oracle network or the \interpreter tree program playing those games we indeed notice that they resepctively fight sharks and fight monkeys which are rewarded by the MDP but that are not the original games goals.

\begin{figure}[h!]
     \centering
     \includegraphics[width=1.\textwidth]{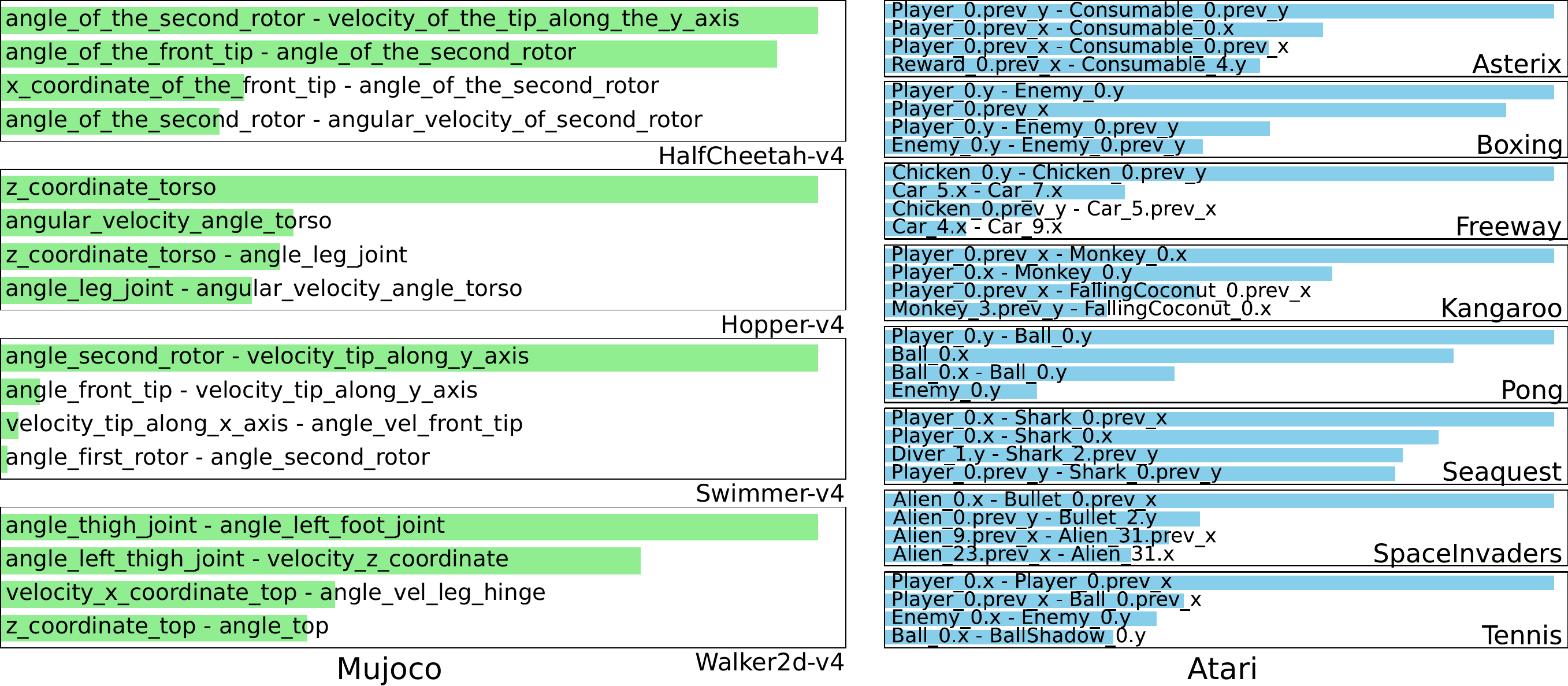}
     \caption{Feature importances on Mujoco (Left) and Atari (Right) environments.}
\end{figure}
%      \begin{subfigure}{0.49\textwidth}
%          \centering
%          \includegraphics[width=1.1\textwidth]{figures/dis_mujoco.pdf}
%      \end{subfigure}
%      \hfill
%      \begin{subfigure}{0.49\textwidth}
%          \centering
%          \includegraphics[width=1.1\textwidth]{figures/fi_atari.pdf}
%      \end{subfigure}
%      \caption{Feature importances}
%     \label{fig:fi-atari-mujoco}
% \end{figure}
\newpage
\section{Extracted programs}
In this section, we show some programs autonomously learned by INTERPRETER.

\begin{figure}[h!]
    \centering
    \tiny
\begin{minted}{Python}
if diver_0.x > 0: # edited
    return "UP"
if Player_0.prev_x - Shark_0.prev_x  <= 0.51:
    if Player_0.prev_y - Shark_0.prev_y  <= -0.56:
        if Shark_0.y - Submarine_1.prev_x  > 1.24 and Player_0.prev_x - Shark_0.x  <= -0.71:
            return "DOWNRIGHTFIRE"
        else:
            return "DOWNFIRE"
    else:
        if Diver_1.prev_y - Shark_2.y  <= 0.92:
            if Player_0.prev_x - Shark_0.x  <= -0.58 or Shark_0.prev_y - Submarine_1.prev_x  <= -3.89:
                return "UPRIGHTFIRE"
            else:
                if Shark_0.y <= 0.93:
                    return "UPFIRE"
                else:
                    return "DOWNFIRE"
        else:
            return "DOWNRIGHTFIRE"
else:
    if Shark_0.y <= 0.53:
        if Player_0.prev_x - Submarine_0.prev_y  <= 0.13:
            return "UPLEFTFIRE"
        else:
            if Diver_1.y - Shark_0.prev_y  <= 1.87:
                if Player_0.prev_y - PlayerMissile_0.prev_y  <= -0.31:
                    return "DOWNFIRE"
                else:
                    return "DOWNLEFT"
            else:
                return "DOWNLEFT"
    else:
        if Shark_0.y - PlayerMissile_0.prev_y  <= 0.18:
            if Player_0.prev_x - Shark_0.x  <= 1.22:
                return "UPFIRE"
            else:
                return "UPLEFTFIRE"
        else:
            return "DOWNLEFTFIRE"
\end{minted}
    \caption{Program to play {Seaquest} returned by \interpreter. The first two lines have been edited by hand to allow the save of divers.}
    \label{fig:seaquest-edited-code}
\end{figure}

\begin{figure}[h!]
    \centering
\begin{minted}{Python}

if Player.y - Ball.y  <= -0.59:
    if Ball.x <= -0.61:
        return "RIGHT"
    else:
        if Player.y - Ball.y  <= -0.81:
            return "LEFT"
        else:
            return "NOOP"
else:
    if Ball.x <= 0.05:
        if Enemy.y <= 0.25:
            return "LEFT"
        else:
            return "RIGHT"
    else:
        if Ball.x - Ball.y > -0.20 and Player.y - Ball.y > -0.43:
            return "RIGHT"
        else:
            return "NOOP"
\end{minted}
    \caption{Program to play Pong returned by \interpreter. Achieving a score of $15.5$.}
    \label{fig:pong-code}
\end{figure}

\section{Program simplification}
\label{app:prog_simpl}
We here show how programs can further be simplified, by merging redundant branches. To reduce the number of nodes/if-else statements in the program, one can for example replace the first block of the following code by the second one:
\begin{minted}{Python}
if Ball.x - Ball.y  <= -0.20:
    return "NOOP"
else:
    if Player.y - Ball.y  <= -0.43:
        return "NOOP"
    else:
        return "UP"

# can be written as

if (Ball.x  -  Ball.y) > -0.20 and (Player.y - Ball.y > -0.43):
        return  "UP"
else:
        return  "NOOP"
\end{minted}

This mainly improves human interpretability, as decisions for specific actions are gathered together.

\section{Related work summary table}
In table \ref{tab:related-work} we summarize existing explainable RL work and compare w.r.t to their required oracle knowledge, their domain ranges, on what problems they were tested, and on the level of explainability they provide. Among the algorithms that require at least an oracle policy, {\small INTERPETER} is the most versatile and well-tested method.
\begin{table}[h!]
\caption{Existing interpretable RL algorithms and their specifications.}
\centering
\begin{tabular}{l|c|c|c|c} 
\toprule
\multicolumn{1}{l|}{Name} & \multicolumn{1}{l|}{oracle Knowledge} & \multicolumn{1}{l|}{$\mathcal{M}$} & \multicolumn{1}{l|}{Benchmarks} & \multicolumn{1}{l}{Programs}\\ \hline
\interpreter & $\pi$ or $Q$   & All & All & Interpretable\\
{\small VIPER} & $\pi$ and $Q$   & $A \in \mathbb{Z}^p$ & Toy & Explainable\\
{\small PIRL} & $\pi$ or $Q$ and Primitives &  All & Toy & Interpretable\\
SCoBots &  $\pi$ or $Q$ and Primitives and LLM & All  & Atari & Interpretable\\
\hline
\hline
{\small NUDGE} &Primitives  & $A \in \mathbb{Z}^p$ & Atari & Interpretable\\
{\small INSIGHT} &  LLM & $A \in \mathbb{Z}^p$ & Atari & Explainable\\
Custard & Primitives & $A \in \mathbb{Z}^p$  & Toy & Interpretable\\
$\pi$-PRL & Primitives & All  & All & Explainable\\
\bottomrule
\end{tabular}
\label{tab:related-work}
\end{table}

\section{User study details}
\label{app:sec:user_study}
Hereafter is provided the detailed questions used during our user studies. We have reached out to AI experts. We have collected and aggregated the answers of $19$ participants. 
\includepdf[pages=-,nup=2x2,landscape=false]{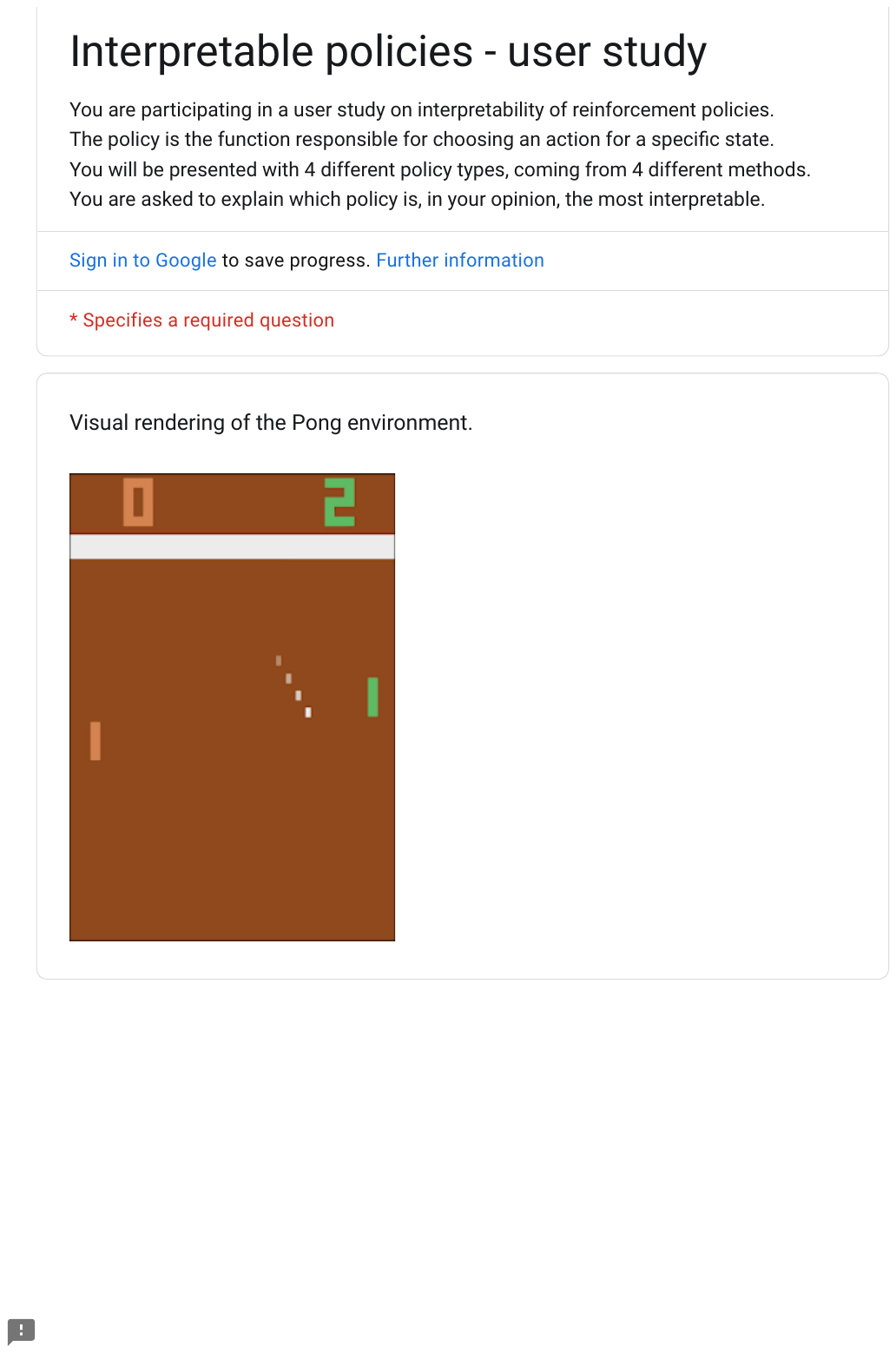}

\newpage
\section*{NeurIPS Paper Checklist}

\begin{enumerate}

\item {\bf Claims}
    \item[] Question: Do the main claims made in the abstract and introduction accurately reflect the paper's contributions and scope?
    \item[] Answer: \answerYes{} % Replace by \answerYes{}, \answerNo{}, or \answerNA{}.
    \item[] Justification: All the algorithms and propositions studied in the intro are studied and presented in detail in the main paper.

\item {\bf Limitations}
    \item[] Question: Does the paper discuss the limitations of the work performed by the authors?
    \item[] Answer: \answerYes{} % Replace by \answerYes{}, \answerNo{}, or \answerNA{}.
    \item[] Justification: Dedicated sections in the experiments and in the conclusion for limitations (section 7).
    \item[] Guidelines:
    \begin{itemize}
        \item The answer NA means that the paper has no limitation while the answer No means that the paper has limitations, but those are not discussed in the paper. 
        \item The authors are encouraged to create a separate "Limitations" section in their paper.
        \item The paper should point out any strong assumptions and how robust the results are to violations of these assumptions (\eg, independence assumptions, noiseless settings, model well-specification, asymptotic approximations only holding locally). The authors should reflect on how these assumptions might be violated in practice and what the implications would be.
        \item The authors should reflect on the scope of the claims made, \eg, if the approach was only tested on a few datasets or with a few runs. In general, empirical results often depend on implicit assumptions, which should be articulated.
        \item The authors should reflect on the factors that influence the performance of the approach. For example, a facial recognition algorithm may perform poorly when image resolution is low or images are taken in low lighting. Or a speech-to-text system might not be used reliably to provide closed captions for online lectures because it fails to handle technical jargon.
        \item The authors should discuss the computational efficiency of the proposed algorithms and how they scale with dataset size.
        \item If applicable, the authors should discuss possible limitations of their approach to address problems of privacy and fairness.
        \item While the authors might fear that complete honesty about limitations might be used by reviewers as grounds for rejection, a worse outcome might be that reviewers discover limitations that aren't acknowledged in the paper. The authors should use their best judgment and recognize that individual actions in favor of transparency play an important role in developing norms that preserve the integrity of the community. Reviewers will be specifically instructed to not penalize honesty concerning limitations.
    \end{itemize}

\item {\bf Theory Assumptions and Proofs}
    \item[] Question: For each theoretical result, does the paper provide the full set of assumptions and a complete (and correct) proof?
    \item[] Answer: \answerNA{} % Replace by \answerYes{}, \answerNo{}, or \answerNA{}.
    \item[] Justification: No theory 

    \item {\bf Experimental Result Reproducibility}
    \item[] Question: Does the paper fully disclose all the information needed to reproduce the main experimental results of the paper to the extent that it affects the main claims and/or conclusions of the paper (regardless of whether the code and data are provided or not)?
    \item[] Answer:  \answerYes{} % Replace by \answerYes{}, \answerNo{}, or \answerNA{}.
    \item[] Justification: Code and data links are provided. Algorithms are described explicitly.
    \item[] Guidelines:
    \begin{itemize}
        \item The answer NA means that the paper does not include experiments.
        \item If the paper includes experiments, a No answer to this question will not be perceived well by the reviewers: Making the paper reproducible is important, regardless of whether the code and data are provided or not.
        \item If the contribution is a dataset and/or model, the authors should describe the steps taken to make their results reproducible or verifiable. 
        \item Depending on the contribution, reproducibility can be accomplished in various ways. For example, if the contribution is a novel architecture, describing the architecture fully might suffice, or if the contribution is a specific model and empirical evaluation, it may be necessary to either make it possible for others to replicate the model with the same dataset, or provide access to the model. In general. releasing code and data is often one good way to accomplish this, but reproducibility can also be provided via detailed instructions for how to replicate the results, access to a hosted model (\eg, in the case of a large language model), releasing of a model checkpoint, or other means that are appropriate to the research performed.
        \item While NeurIPS does not require releasing code, the conference does require all submissions to provide some reasonable avenue for reproducibility, which may depend on the nature of the contribution. For example
        \begin{enumerate}
            \item If the contribution is primarily a new algorithm, the paper should make it clear how to reproduce that algorithm.
            \item If the contribution is primarily a new model architecture, the paper should describe the architecture clearly and fully.
            \item If the contribution is a new model (\eg, a large language model), then there should either be a way to access this model for reproducing the results or a way to reproduce the model (\eg, with an open-source dataset or instructions for how to construct the dataset).
            \item We recognize that reproducibility may be tricky in some cases, in which case authors are welcome to describe the particular way they provide for reproducibility. In the case of closed-source models, it may be that access to the model is limited in some way (\eg, to registered users), but it should be possible for other researchers to have some path to reproducing or verifying the results.
        \end{enumerate}
    \end{itemize}

\item {\bf Open access to data and code}
    \item[] Question: Does the paper provide open access to the data and code, with sufficient instructions to faithfully reproduce the main experimental results, as described in supplemental material?
    \item[] Answer: \answerYes{} % Replace by \answerYes{}, \answerNo{}, or \answerNA{}.
    \item[] Justification: Anonymized github repo and data links are provided.
    \item[] Guidelines:
    \begin{itemize}
        \item The answer NA means that paper does not include experiments requiring code.
        \item Please see the NeurIPS code and data submission guidelines (\url{https://nips.cc/public/guides/CodeSubmissionPolicy}) for more details.
        \item While we encourage the release of code and data, we understand that this might not be possible, so “No” is an acceptable answer. Papers cannot be rejected simply for not including code, unless this is central to the contribution (\eg, for a new open-source benchmark).
        \item The instructions should contain the exact command and environment needed to run to reproduce the results. See the NeurIPS code and data submission guidelines (\url{https://nips.cc/public/guides/CodeSubmissionPolicy}) for more details.
        \item The authors should provide instructions on data access and preparation, including how to access the raw data, preprocessed data, intermediate data, and generated data, etc.
        \item The authors should provide scripts to reproduce all experimental results for the new proposed method and baselines. If only a subset of experiments are reproducible, they should state which ones are omitted from the script and why.
        \item At submission time, to preserve anonymity, the authors should release anonymized versions (if applicable).
        \item Providing as much information as possible in supplemental material (appended to the paper) is recommended, but including URLs to data and code is permitted.
    \end{itemize}

\item {\bf Experimental Setting/Details}
    \item[] Question: Does the paper specify all the training and test details (\eg, data splits, hyperparameters, how they were chosen, type of optimizer, etc.) necessary to understand the results?
    \item[] Answer: \answerYes{} % Replace by \answerYes{}, \answerNo{}, or \answerNA{}.
    \item[] Justification: Evertyhing is detailed clearly in the main paper, in the appendix and in the code repos for baselines and contributed algorithm.

\item {\bf Experiment Statistical Significance}
    \item[] Question: Does the paper report error bars suitably and correctly defined or other appropriate information about the statistical significance of the experiments?
    \item[] Answer:  \answerYes{} %, \answerNo{}, or \answerNA{}.
    \item[] Justification: When processes are stochastic error bars are plotted or written.
    \item[] Guidelines:
    \begin{itemize}
        \item The answer NA means that the paper does not include experiments.
        \item The authors should answer "Yes" if the results are accompanied by error bars, confidence intervals, or statistical significance tests, at least for the experiments that support the main claims of the paper.
        \item The factors of variability that the error bars are capturing should be clearly stated (for example, train/test split, initialization, random drawing of some parameter, or overall run with given experimental conditions).
        \item The method for calculating the error bars should be explained (closed form formula, call to a library function, bootstrap, etc.)
        \item The assumptions made should be given (\eg, Normally distributed errors).
        \item It should be clear whether the error bar is the standard deviation or the standard error of the mean.
        \item It is OK to report 1-sigma error bars, but one should state it. The authors should preferably report a 2-sigma error bar than state that they have a 96\% CI, if the hypothesis of Normality of errors is not verified.
        \item For asymmetric distributions, the authors should be careful not to show in tables or figures symmetric error bars that would yield results that are out of range (\eg negative error rates).
        \item If error bars are reported in tables or plots, The authors should explain in the text how they were calculated and reference the corresponding figures or tables in the text.
    \end{itemize}

\item {\bf Experiments Compute Resources}
    \item[] Question: For each experiment, does the paper provide sufficient information on the computer resources (type of compute workers, memory, time of execution) needed to reproduce the experiments?
    \item[] Answer: \answerYes{} % Replace by \answerYes{}, \answerNo{}, or \answerNA{}.
    \item[] Justification: Exact CPU model as well as ram are provided.
    \item[] Guidelines:
    \begin{itemize}
        \item The answer NA means that the paper does not include experiments.
        \item The paper should indicate the type of compute workers CPU or GPU, internal cluster, or cloud provider, including relevant memory and storage.
        \item The paper should provide the amount of compute required for each of the individual experimental runs as well as estimate the total compute. 
        \item The paper should disclose whether the full research project required more compute than the experiments reported in the paper (\eg, preliminary or failed experiments that didn't make it into the paper). 
    \end{itemize}
    
\item {\bf Code Of Ethics}
    \item[] Question: Does the research conducted in the paper conform, in every respect, with the NeurIPS Code of Ethics \url{https://neurips.cc/public/EthicsGuidelines}?
    \item[] Answer: \answerYes{} %, \answerNo{}, or \answerNA{}.
    \item[] Justification: Research is done ethically with a lot of concerns for reproduciblity and validity of the results.
    \item[] Guidelines:
    \begin{itemize}
        \item The answer NA means that the authors have not reviewed the NeurIPS Code of Ethics.
        \item If the authors answer No, they should explain the special circumstances that require a deviation from the Code of Ethics.
        \item The authors should make sure to preserve anonymity (\eg, if there is a special consideration due to laws or regulations in their jurisdiction).
    \end{itemize}

\item {\bf Broader Impacts}
    \item[] Question: Does the paper discuss both potential positive societal impacts and negative societal impacts of the work performed?
    \item[] Answer: \answerNA{} % Replace by \answerYes{}, \answerNo{}, or \answerNA{}.
    \item[] Justification: No societal impact, our work simply proposes a distillation algorithms applied on widely known tasks.
    \item[] Guidelines:
    \begin{itemize}
        \item The answer NA means that there is no societal impact of the work performed.
        \item If the authors answer NA or No, they should explain why their work has no societal impact or why the paper does not address societal impact.
        \item Examples of negative societal impacts include potential malicious or unintended uses (\eg, disinformation, generating fake profiles, surveillance), fairness considerations (\eg, deployment of technologies that could make decisions that unfairly impact specific groups), privacy considerations, and security considerations.
        \item The conference expects that many papers will be foundational research and not tied to particular applications, let alone deployments. However, if there is a direct path to any negative applications, the authors should point it out. For example, it is legitimate to point out that an improvement in the quality of generative models could be used to generate deepfakes for disinformation. On the other hand, it is not needed to point out that a generic algorithm for optimizing neural networks could enable people to train models that generate Deepfakes faster.
        \item The authors should consider possible harms that could arise when the technology is being used as intended and functioning correctly, harms that could arise when the technology is being used as intended but gives incorrect results, and harms following from (intentional or unintentional) misuse of the technology.
        \item If there are negative societal impacts, the authors could also discuss possible mitigation strategies (\eg, gated release of models, providing defenses in addition to attacks, mechanisms for monitoring misuse, mechanisms to monitor how a system learns from feedback over time, improving the efficiency and accessibility of ML).
    \end{itemize}
    
\item {\bf Safeguards}
    \item[] Question: Does the paper describe safeguards that have been put in place for responsible release of data or models that have a high risk for misuse (\eg, pretrained language models, image generators, or scraped datasets)?
    \item[] Answer: \answerNA{} % Replace by \answerYes{}, \answerNo{}, or \answerNA{}.
    \item[] Justification: no risk (see above)

\item {\bf Licenses for existing assets}
    \item[] Question: Are the creators or original owners of assets (\eg, code, data, models), used in the paper, properly credited and are the license and terms of use explicitly mentioned and properly respected?
    \item[] Answer: \answerYes{} % Replace by \answerYes{}, \answerNo{}, or \answerNA{}.
    \item[] Justification: Appropriate credits is given when necessary and other code not from the authors are open sourced.
    \item[] Guidelines:
    \begin{itemize}
        \item The answer NA means that the paper does not use existing assets.
        \item The authors should cite the original paper that produced the code package or dataset.
        \item The authors should state which version of the asset is used and, if possible, include a URL.
        \item The name of the license (\eg, CC-BY 4.0) should be included for each asset.
        \item For scraped data from a particular source (\eg, website), the copyright and terms of service of that source should be provided.
        \item If assets are released, the license, copyright information, and terms of use in the package should be provided. For popular datasets, \url{paperswithcode.com/datasets} has curated licenses for some datasets. Their licensing guide can help determine the license of a dataset.
        \item For existing datasets that are re-packaged, both the original license and the license of the derived asset (if it has changed) should be provided.
        \item If this information is not available online, the authors are encouraged to reach out to the asset's creators.
    \end{itemize}

\item {\bf New Assets}
    \item[] Question: Are new assets introduced in the paper well documented and is the documentation provided alongside the assets?
    \item[] Answer: \answerYes{} % Replace by \answerYes{}, \answerNo{}, or \answerNA{}.
    \item[] Justification: Code is documented to the best we could.

\item {\bf Crowdsourcing and Research with Human Subjects}
    \item[] Question: For crowdsourcing experiments and research with human subjects, does the paper include the full text of instructions given to participants and screenshots, if applicable, as well as details about compensation (if any)? 
    \item[] Answer:\answerYes{} % Replace by \answerYes{}, \answerNo{}, or \answerNA{}.
    \item[] Justification: PDF of the user study questions are provided in the Appendix H.

\item {\bf Institutional Review Board (IRB) Approvals or Equivalent for Research with Human Subjects}
    \item[] Question: Does the paper describe potential risks incurred by study participants, whether such risks were disclosed to the subjects, and whether Institutional Review Board (IRB) approvals (or an equivalent approval/review based on the requirements of your country or institution) were obtained?
    \item[] Answer: \answerNA{} % Replace by \answerYes{}, \answerNo{}, or \answerNA{}.
    \item[] Justification: the user study is more of a proof of concetps than a large scale study with risks.

\end{enumerate}

\end{document}